\documentclass[journal,10pt,twocolumn]{IEEETran}
\usepackage[utf8]{inputenc}

\usepackage{threeparttable}
\usepackage{cite}
\usepackage{amsmath,amssymb,amsfonts}
\usepackage{algorithmic}
\usepackage{graphicx}
\usepackage{textcomp}
\usepackage{indentfirst}
\usepackage{graphicx}
\usepackage{csquotes}
\usepackage{multirow}
\usepackage{tabularx}
\usepackage{float}
\usepackage{color}
\usepackage{framed}
\usepackage{comment}
\usepackage{caption3}
\usepackage[boxed]{algorithm2e}

\newcommand{\todob}[1]{{\color{black} #1}}
\renewcommand{\figurename}{Figure}
\def\tablename{Table}
\begin{document}

\title{Comprehensive SNN Compression Using ADMM Optimization and Activity Regularization}

\author{Lei Deng, \IEEEmembership{Member}, \IEEEmembership{IEEE}, Yujie Wu, Yifan Hu, Ling Liang, Guoqi Li, \IEEEmembership{Member}, \IEEEmembership{IEEE}, Xing Hu, \\Yufei Ding, Peng Li, \IEEEmembership{Fellow}, \IEEEmembership{IEEE}, Yuan Xie, \IEEEmembership{Fellow}, \IEEEmembership{IEEE} \vspace{-20pt}\\

\thanks{The work was partially supported by National Science Foundation (Grant No. 1725447), Tsinghua University Initiative Scientific Research Program, Tsinghua-Foshan Innovation Special Fund (TFISF), and National Natural Science Foundation of China (Grant No. 61876215). Lei Deng and Yujie Wu contributed equally to this work, corresponding authors: Guoqi Li and Xing Hu. Lei Deng is with the Center for Brain Inspired Computing Research, Department of Precision Instrument, Tsinghua University, Beijing 100084, China, and also with the Department of Electrical and Computer Engineering, University of California, Santa Barbara, CA 93106, USA (email: leideng@ucsb.edu). Yujie Wu, Yifan Hu, and Guoqi Li are with the Center for Brain Inspired Computing Research, Department of Precision Instrument, Tsinghua University, Beijing 100084, China (email: \{wu-yj16, huyf19\}@mails.tsinghua.edu.cn, liguoqi@mail.tsinghua.edu.cn). Ling Liang, Peng Li, and Yuan Xie are with the Department of Electrical and Computer Engineering, University of California, Santa Barbara, CA 93106, USA (email: \{lingliang, lip, yuanxie\}@ucsb.edu). Xing Hu is with the State Key Laboratory of Computer Architecture, Institute of Computing Technology, Chinese Academy of Sciences, Beijing 100190, China (email: huxing@ict.ac.cn). Yufei Ding is with the Department of Computer Science, University of California, Santa Barbara, CA 93106, USA (email: yufeiding@cs.ucsb.edu). }}

\maketitle

\begin{abstract}
 
As well known, the huge memory and compute costs of both artificial neural networks (ANNs) and spiking neural networks (SNNs) greatly hinder their deployment on edge devices with high efficiency. Model compression has been proposed as a promising technique to improve the running efficiency via parameter and operation reduction. Whereas, this technique is mainly practiced in ANNs rather than SNNs. It is interesting to answer how much an SNN model can be compressed without compromising its functionality, where two challenges should be addressed: i) the accuracy of SNNs is usually sensitive to model compression, which requires an accurate compression methodology; ii) the computation of SNNs is event-driven rather than static, which produces an extra compression dimension on dynamic spikes. To this end, we realize a comprehensive SNN compression through three steps. First, we formulate the connection pruning and weight quantization as a constrained optimization problem. Second, we combine spatio-temporal backpropagation (STBP) and alternating direction method of multipliers (ADMM) to solve the problem with minimum accuracy loss. Third, we further propose activity regularization to reduce the spike events for fewer active operations. These methods can be applied in either a single way for moderate compression or a joint way for aggressive compression. We define several quantitative metrics to evaluation the compression performance for SNNs. Our methodology is validated in pattern recognition tasks over MNIST, N-MNIST, CIFAR10, and CIFAR100 datasets, where extensive comparisons, analyses, and insights are provided. To our best knowledge, this is the first work that studies SNN compression in a comprehensive manner by exploiting all compressible components and achieves better results.

\end{abstract}


\section{Introduction}\label{sec:intro}

Neural networks, constructed by a plenty of nodes (neurons) and connections (synapses), are powerful in information representation, which has been evidenced in a wide spectrum of intelligent tasks such as visual or auditory recognition \cite{he2016deep, abdel2014convolutional, esser2016convolutional, diehl2016truehappiness}, language modelling \cite{vaswani2017attention, bellec2018long}, medical diagnosis \cite{esteva2017dermatologist, kasabov2015spiking}, game playing \cite{silver2016mastering}, heuristic
solution of hard computational problems \cite{maass2014noise}, sparse coding \cite{knag2015sparse}, etc. The models include two categories: application-oriented artificial neural networks (ANNs) and neuroscience-oriented spiking neural networks (SNNs). The former process continuous signals layer by layer with nonlinear activation functions; while the latter integrate temporal information via neuronal dynamics and use binary spike signals (0-nothing or 1-spike event) for inter-neuron communication. The success of these models spurs numerous researchers to study domain-specific hardwares for ANNs and SNNs, termed as deep learning accelerators \cite{chen2014dadiannao, chen2017eyeriss, jouppi2017datacenter} and neuromorphic chips \cite{furber2014spinnaker, merolla2014million, davies2018loihi}, respectively.

Whereas, the huge amount of parameters and operations in neural networks greatly limits the running performance and hinders the deployment on edge devices with tight resources. To solve this problem, various model compression technologies including low-rank decomposition \cite{novikov2015tensorizing}, network sparsification \cite{han2015learning, wen2016learning, he2018amc, zhang2018systematic}, and data quantization \cite{courbariaux2016binarized, zhou2016dorefa, deng2018gxnor, wang2019haq} have been proposed to shrink the model size, which is quite helpful in boosting the hardware performance \cite{huang2018highly, zhang2016cambricon, aimar2018nullhop, liang2018crossbar, han2016eie, lee2018unpu, andri2018yodann}. Although this solution has become a promising way to reduce the memory and compute costs in deep learning, it has yet to be well studied in the neuromorphic computing domain. The underlying reason is because the behaviors of SNNs are quite different from those of ANNs. For example, i) the spike coding of SNNs makes the accuracy very sensitive to  model compression , which demands an accurate compression methodology; ii) the processing of SNNs is event-driven with a dynamic rather than static execution pattern, which produces an extra compression dimension on dynamic spikes.

In fact, we find several previous work that tried tentative explorations on the SNN compression topic. A two-stage growing-pruning algorithm for compact fully-connected (FC) SNNs was verified on small-scale datasets \cite{dora2015two}. Based on a single FC layer with spike timing dependent plasticity (STDP) learning rule, a soft-pruning method (setting part of weights to a lower bound during training) achieved 95.04\% accuracy on MNIST \cite{shi2019soft}. Similarly on FC-based SNNs with STDP, both connection pruning and weight quantization were conducted and validated on MNIST with 91.5\% accuracy \cite{rathi2018stdp}. Combining an FC feature extraction layer with binary weights trained by stochastic STDP and an FC classification layer with 24-bit precision, A. Yousefzadeh et al. \cite{yousefzadeh2018practical} presented 95.7\% accuracy on MNIST. S. K. Esser et al. \cite{esser2016convolutional} adapted normal ANN models to their variants with ternary weights and binary activations, and then deployed them on the TrueNorth chip that only supports SNNs. B. Rueckauer et al.\cite{Bodo2017Conversion} converted the binarized ANN models to their SNN counterparts and then analyzed the accuracy-vs.-operations trade-off. However, the techniques with adaption from ANNs suffers from costly computation in the ANN domain and conversion between ANN and SNN, even though we only expect the resulting compressed SNN model.  G. Srinivasan et al. \cite{srinivasan2019restocnet} introduced residual paths into SNNs and combined spiking convolutional (Conv) layers with binary weight kernels trained by probabilistic STDP and non-spiking FC layers trained by conventional backpropagation (BP) algorithm, which demonstrated 98.54\% accuracy on MNIST but only 66.23\% accuracy on CIFAR10. Unfortunately, these existing works on SNN compression did not either harness large-scale models with impressive performance or touch  normal  SNNs (just ANN variants,  not straightfoward enough for the SNN compression ).

Hence, we formally raise a question that \emph{how much an SNN model can be compressed without compromising much functionality}. We answer this question through three steps. (1) First, we formulate the connection pruning and the weight quantization as a constrained optimization problem based on supervised learning. (2) Second, we combine the emerging spatio-temporal backpropagation (STBP) supervised learning \cite{wu2018spatio, wu2019direct} and the powerful alternating direction method of multipliers (ADMM) optimization tool \cite{boyd2011distributed} to solve the problem with minimum accuracy loss. (3) Third, we propose activity regularization to reduce the number of spike events, leading to fewer active operations. These approaches can be flexibly used in a single or joint manner according to actual needs for compression performance. We comprehensively validate our methods in SNN-based pattern recognition tasks over MNIST, N-MNIST, CIFAR10, and CIFAR100 datasets. Several quantitative metrics to evaluate the compression ratio are defined, based on which a variety of comparisons between different compression strategies and in-depth result analyses are conducted. Our work can achieve aggressive compression ratio with advanced accuracy maintaining, which promises ultra-efficient neuromorphic systems.

For better readability, we briefly summarize our contributions as follows:
\begin{itemize}
\item We present the first work that investigates comprehensive and aggressive compression for SNNs by exploiting all compressible components  and defining quantitative evaluation metrics.

\item The effectiveness of the ADMM optimization tool is validated on SNNs to reduce the parameter memory space and baseline compute cost for the first time. Then, the activity regularization method is further proposed to reduce the number of active operations. All the proposed approaches can be flexibly applied in either a single way for moderate compression or a joint way for aggressive compression.

\item We demonstrate high compression performance in SNN-based pattern recognition tasks with acceptable accuracy degradation. Rich contrast experiments, in-depth result analyses, and interesting insights are provided. 

\end{itemize}

The rest of this paper is organized as follows: Section \ref{sec:preliminary} introduces some preliminaries of the SNN model, the STBP learning algorithm, and the ADMM optimization approach; Section \ref{sec:approach} systematically explains the possible compression ways, the proposed ADMM-based connection pruning and weight quantization, the activity regularization, their joint use, and the evaluation metrics; The experimental setup, experimental results, and in-depth analyses are provided in Section \ref{sec:results}; Finally, Section \ref{sec:conclusion} concludes and discusses the paper.
\section{Preliminaries}\label{sec:preliminary}

\subsection{Spiking Neural Networks}

\begin{figure}[htbp]
\centering
\includegraphics[width=0.4\textwidth]{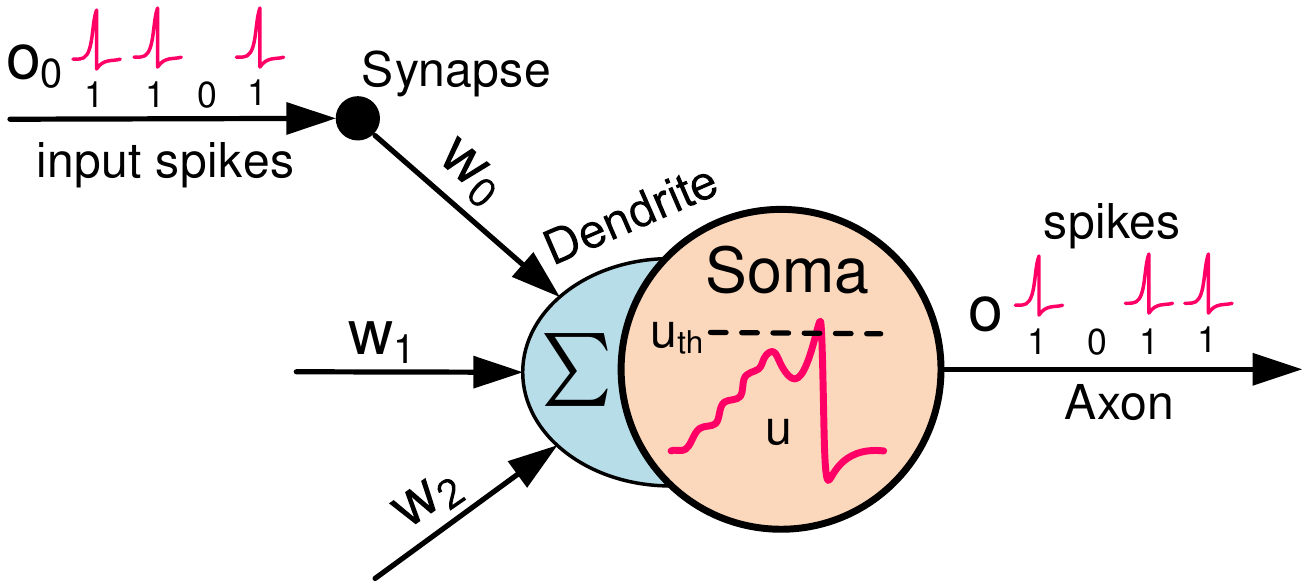}
\caption{Illustration of a spiking neuron comprised of synapses, dendrites, soma, and axon.}
\label{fig:snn_neuron}
\end{figure}

In a neural network, neurons behave as the basic processing units which are wired by abundant synapses. Each synapse has a weight that affects the signal transfer efficacy. Figure \ref{fig:snn_neuron} presents a typical spiking neuron, which is comprised of synapses, dendrites, soma, and axon. Dendrites integrate the weighted input spikes and the soma consequently conducts nonlinear transformation to produce output spikes, then the axon transfers these output spikes to post-neurons. The neuronal behaviors can be described by the classic leaky integrate-and-fire (LIF) model \cite{gerstner2014neuronal} as follows:
\begin{equation}
\begin{cases}
\tau\frac{du(t)}{dt}=-[u(t)-u_{r_1}]+\sum_j w_{j} \sum_{t^k_j\in [t-T_w,~t]}K(t-t^k_j)\\
\begin{cases}
o(t)=1~\&~u(t)=u_{r_2},~\text{if}~u(t)\geq u_{th}\\
o(t)=0,~\text{if}~u(t)< u_{th}~
\end{cases}
\end{cases}
\label{equ:snn_neuron}
\end{equation}
where $(t)$ denotes the timestep, $\tau$ is a time constant, $u$ is the membrane potential of current neuron, and $o$ is the output spike event. $w_{j}$ is the synaptic weight from the $j$-th input neuron to the current neuron, and $t^k_j$ is the timestep when the $k$-th spike from the $j$-th input neuron comes during the past integration time window of $T_w$. $K(\cdot)$ is a kernel function describing the temporal decay effect that a more recent spike should have a greater impact on the post-synaptic membrane potential. $u_{r_1}$ and $u_{r_2}$ are the resting potential and reset potential, respectively, and $u_{th}$ is a threshold that determines whether to fire a spike or not.

According to Equation (\ref{equ:snn_neuron}), SNNs have the following differences compared with ANNs: (1) each neuron has temporal dynamics, i.e. memorization of the historical states; (2) the multiplication operations during integration can be removed when $T_w=1$ owing to the binary spike inputs; (3) the network activities are very sparse because each neuron remains silent if the membrane potential does not exceed the firing threshold. In summary, the temporal memorization makes it well-suited for dynamic data with timing information, and the spike-driven paradigm with sparse activities enables power-efficient asynchronous circuit design.

\subsection{STBP Supervised Learning}

There exist three categories of learning algorithms for SNNs: unsupervised \cite{diehl2015unsupervised, mozafari2018combining}, indirectly supervised \cite{diehl2015fast, sengupta2019going, hu2018spiking}, and directly supervised \cite{lee2016training, wu2018spatio, jin2018hybrid, wu2019direct}.  Note that here the ``indirectly supervised'' mainly refers to the ANN-to-SNN-conversion learning. It adopts supervised learning during training, whereas, the supervised learning is applied on the ANN model rather than the SNN one converted from the ANN during inference. In contrast, the supervised learning is directly applied on the SNN model in ``directly supervised'' algorithms.  Since SNN compression requires an accurate learning method and the ADMM optimization (to be shown latter) relies on the supervised learning framework, we select an emerging directly supervised training algorithm, named spatio-temporal backpropagation (STBP) \cite{wu2018spatio, wu2019direct}. We do not use the indirectly supervised training due to the complex model transformation between ANNs and SNNs. 

STBP is based on an iterative version of the LIF model in Equation (\ref{equ:snn_neuron}). Specifically, it yields
\begin{equation}
	\begin{cases}
	\label{equ:lif_iterative}
	u^{t+1,n+1}_i &= e^{-\frac{dt}{\tau}}u^{t,n+1}_i(1 - o^{t,n+1}_i)+
	\sum_j w_{ij}^n o^{t+1,n}_j \\
	o^{t+1,n+1}_i &= H(u^{t+1,n+1}_i - u_{th})
	\end{cases}
\end{equation}	
where $dt$ is the length of the simulation timestep, $o$ denotes the neuronal spike output, $t$ and $n$ are indices of timestep and layer, respectively. $e^{-\frac{dt}{\tau}}$ reflects the leakage effect of the membrane potential.  $H(\cdot)$ is the Heaviside step function, i.e., $H(x)=1$ when $x\geq0$; $H(x)=0$ otherwise.  This iterative LIF format incorporates all behaviors including integration, fire, and reset in the original neuron model. For simplicity, here we set the parameters in Equation (\ref{equ:snn_neuron}) with $u_{r_1}=u_{r_2}=0$, $T_w=1$, and $K(\cdot)\equiv 1$. 

STBP uses rate coding to represent information, wherein the number of spikes matters. The loss function is given by
\begin{equation}
\label{equ:stbp_loss}
L = \parallel  \pmb{Y}^{label} - \frac{1}{T} \sum_{t=1}^T  \pmb{O}^{t,N} \parallel_2^2.
\end{equation}
This loss function measures the discrepancy between the ground truth and the firing rate of the output layer (i.e. the $N$-th layer) during the given simulation time window $T$.  In fact, Equation (\ref{equ:stbp_loss}) reflects how to determine the recognition accuracy: i) each output neuron integrates the spikes along all the $T$ timesteps and normalizes the result by dividing $T$ to get a normalized average fire rate value within $[0,~1]$; ii) the output neuron with the largest average fire rate corresponds to the recognized class. 

Given Equation (\ref{equ:lif_iterative})-(\ref{equ:stbp_loss}), the gradient propagation and parameter update in STBP can be derived as follows 
\begin{equation}
\begin{cases}
\label{equ:stbp}
\frac{\partial L}{\partial o_i^{t,n}}
=
\sum_{j}\frac{\partial L}{\partial u_j^{t,n+1}}\frac{\partial u_j^{t,n+1}}{\partial o_i^{t,n}} + \frac{\partial L}{\partial u_i^{t+1,n}}\frac{\partial u_i^{t+1,n}}{\partial o_i^{t,n}} ,
\\
\frac{\partial L}{\partial u_i^{t,n}}
=
\frac{\partial L}{\partial o_i^{t,n}} \frac{\partial o_i^{t,n}}{\partial u_i^{t,n}}+\frac{\partial L}{\partial u_i^{t+1,n}}\frac{\partial u_i^{t+1,n}}{\partial u_i^{t,n}},\\
\bigtriangledown  {w^n_{ji}} 
= 
\sum_{t=1}^{T}\frac{\partial L}{\partial  {u_j^{t,n+1}}}{ {o_i^{t,n}}}.
\end{cases}
\end{equation}
The derivative approximation method can be used to calculate $\frac{\partial o}{\partial u}$ \cite{wu2018spatio}. Specifically, it is governed by  $\frac{\partial o}{\partial u} = H' \approx boxcar(u_{th}-\frac{a}{2},~u_{th}+\frac{a}{2};~u)$ .  Note that $boxcar(u_{th}-\frac{a}{2},~u_{th}+\frac{a}{2};~u)$ is the $boxcar$ function defined by the sum of two Heaviside step functions, i.e., $\frac{1}{a}\{H[u-(u_{th}-\frac{a}{2})] - H[u - (u_{th}+\frac{a}{2})]\}$ , where $a$ is a hyper-parameter that determines the gradient width.

\subsection{ADMM Optimization Tool}

ADMM is a classic and powerful tool to solve constrained optimization problems \cite{boyd2011distributed}. The main idea of ADMM is to decompose the original non-differentiable optimization problem to a differentiable sub-problem which can be solved by gradient descent and a non-differentiable sub-problem with an analytical or heuristic solution. 

The basic problem of ADMM can be described as
\begin{equation}
\underset{\pmb{X},~\pmb{Z}}{\text{min}} ~f(\pmb{X})+g(\pmb{Z}),~s.t.~A\pmb{X}+B\pmb{Z}=\pmb{C}
\label{equ:admm1}
\end{equation}
where we assume $\pmb{X}\in R^{N}$, $\pmb{Z}\in R^{M}$, $A\in R^{K\times N}$, $B\in R^{K\times M}$, $\pmb{C}\in R^{K}$. $f(\cdot)$ is the major cost function which is usually differentiable and $g(\cdot)$ is an indicator of constraints which is usually non-differentiable. Then, the greedy optimization of its augmented Lagrangian \cite{boyd2011distributed}, $L_\rho(\pmb{X},~\pmb{Z},~\pmb{Y})=f(\pmb{X})+g(\pmb{Z})+\pmb{Y}^T(A\pmb{X}+B\pmb{Z}-\pmb{C})+\frac{\rho}{2}\|A\pmb{X}+B\pmb{Z}-\pmb{C}\|_2^2$, can be iteratively calculated by
\begin{equation}
\begin{cases}
\pmb{X}^{n+1}=\underset{\pmb{X}}{\text{arg min}}~L_\rho(\pmb{X},~\pmb{Z}^n,~\pmb{Y}^n)\\
\pmb{Z}^{n+1}=\underset{\pmb{Z}}{\text{arg min}}~L_\rho(\pmb{X}^{n+1},~\pmb{Z},~\pmb{Y}^n)\\
\pmb{Y}^{n+1}=\pmb{Y}^{n}+\rho(A\pmb{X}^{n+1}+B\pmb{Z}^{n+1}-\pmb{C})
\end{cases}
\label{equ:admm2}
\end{equation}
where $\pmb{Y}$ is the Lagrangian multipliers and $\rho$ is a penalty coefficient. The $\pmb{X}$ minimization sub-problem is differentiable that is easy to solve via gradient descent. The $\pmb{Z}$ minimization sub-problem is non-differentiable, but fortunately it can usually be solved analytically or heuristically.
\section{Spiking Neural Network Compression}\label{sec:approach}

In this section, we first give the possible compression ways, and then explain the proposed compression approaches, algorithms, and evaluation metrics in detail.

\begin{figure*}[!htbp]
\centering
\includegraphics[width=0.72\textwidth]{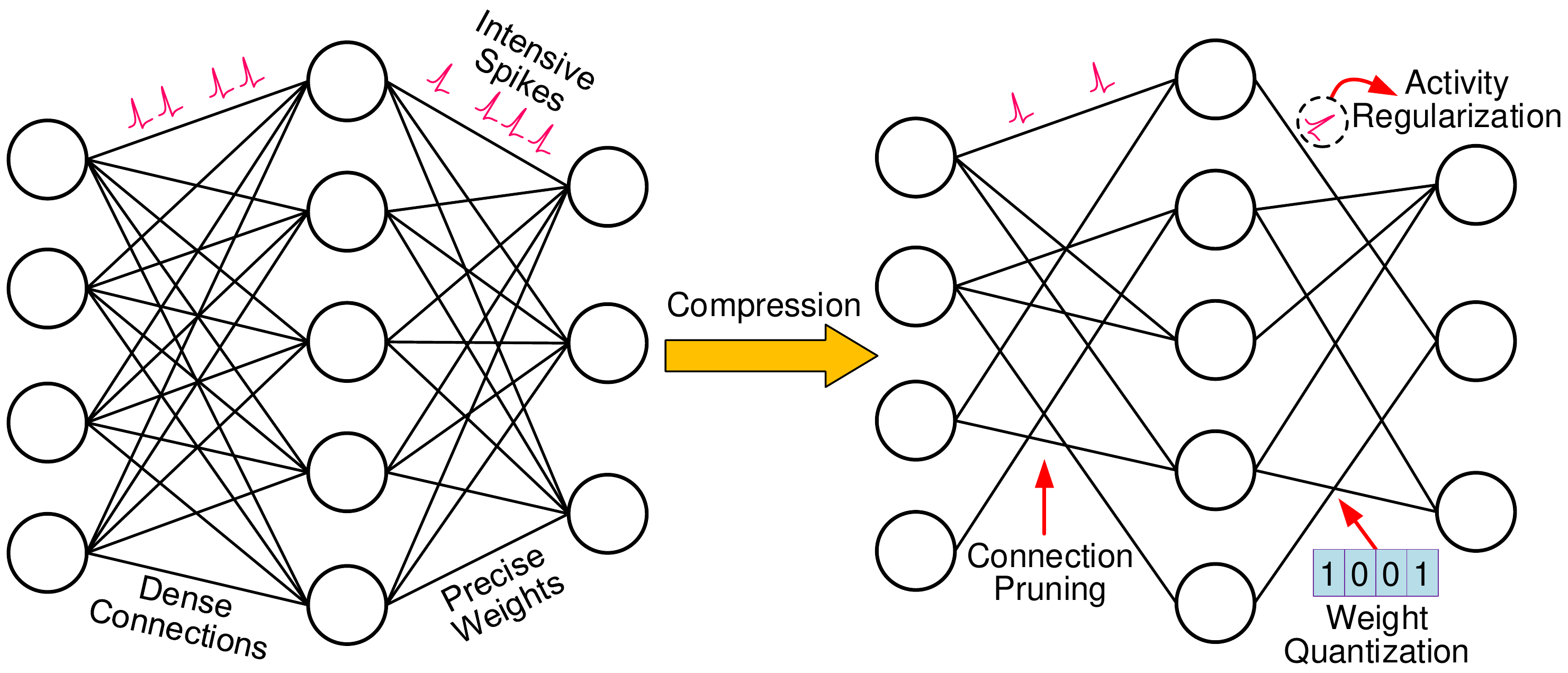}
\caption{Possible ways for SNN inference compression: connection pruning and weight quantization for memory saving and baseline operation reduction, and activity regularization for active operation reduction. }
\label{fig:snn_compression}
\end{figure*}

\subsection{Possible Compression Ways}\label{sec:compression_way}

The compression of SNNs in this work targets the reduction of memory and computation in inference. Figure \ref{fig:snn_compression} illustrates the possible ways to compress an SNN model. On the memory side, synapses occupy the most storage space. There are usually two ways to reduce the synapse memory: the number of connections and the bitwidth of weights. On the compute side, although the connection pruning and the weight quantization already help reduce the amount of operations, there is an additional compression way on the dynamic spikes. As well known, the total number of operations for an SNN layer can be governed by $N_{ops}\cdot R$ \cite{deng2020rethinking}, where $N_{ops}$ is the number of baseline operations and $R\in [0,~1]$ is the average spike rate per neuron per timestep that usually determines the active power of neuromorphic chips \cite{merolla2014million}.

To realize a comprehensive SNN compression considering all the above ways, we first try to combine the STBP supervised learning and the ADMM optimization tool for connection pruning and weight quantization to shrink memory and reduce $N_{ops}$. The reason that we combine STBP and ADMM is two-fold: (1) ADMM recently shows an impressive compression ratio with good accuracy maintaining in the ANN domain \cite{leng2018extremely, zhang2018systematic, zhang2018structadmm, ye2018progressive, ren2019admm}; (2) ADMM requires a supervised learning framework, which excludes the conventional unsupervised learning algorithms for SNNs. Then, besides synapse compression, we additionally propose an activity regularization to reduce $R$ for a further reduction of operations. We will explain these methods one by one in the rest subsections.

\begin{algorithm}
\label{alg:admm_prune}
\caption{ADMM-based Connection Pruning}

\KwData{$s$ (connection sparsity)}

\begin{center}\textbf{Step I: ADMM Retraining for Pruning}\\ \end{center}
Initialize $\pmb{W}^0$ with the pre-trained weights;\\
Initialize $\widetilde{\pmb{Y}}^0=\pmb{0}$;\\
Initialize $\pmb{Z}_p^0$ with $\pmb{W}^0$ and zero out the $s\%$ magnitude-smallest elements;\\

\KwData{$\pmb{W}^n$, $\pmb{Z}_p^n$, and $\widetilde{\pmb{Y}}^n$ after the $n$-th iteration}
\KwResult{$\pmb{W}^{n+1}$, $\pmb{Z}_p^{n+1}$, and $\widetilde{\pmb{Y}}^{n+1}$}

1. Rewrite the loss function:\\
~~~$L\Leftarrow f(\pmb{W})+\frac{\rho}{2}\|\pmb{W}-\pmb{Z}_p^{n}+\widetilde{\pmb{Y}}^n\|_2^2$;\\
2. Update weights:\\
~~~$\pmb{W}^{n+1}\Leftarrow$ retrain the SNN model one more iteration;\\
3. Update $\pmb{Z}_p^{n+1}$:\\
~~~$\pmb{Z}_p^{n+1}\Leftarrow$ [zero out the $s$ fraction of magnitude-smallest elements in $(\pmb{W}^{n+1}+\widetilde{\pmb{Y}}^n)$];\\
4. Update $\widetilde{\pmb{Y}}^{n+1}$:\\
~~~$\widetilde{\pmb{Y}}^{n+1}\Leftarrow (\widetilde{\pmb{Y}}^n+\pmb{W}^{n+1}-\pmb{Z}_p^{n+1})$;\\

~\\
\begin{center}\textbf{Step II: Hard-Pruning Retraining}\\ \end{center}
Initialize $\pmb{W}_p^0$ with the weights from Step I;\\
Initialize the loss function $L=f(\pmb{W}_p)$;\\

\KwData{$\pmb{W}_p^n$ after the $n$-th iteration}
\KwResult{$\pmb{W}_p^{n+1}$}

Update weights:\\
~~~$\pmb{W}_p^{n+1}\Leftarrow$ retrain the SNN model one more iteration;\\
~~~$\pmb{W}_p^{n+1}\Leftarrow$ [zero out the $s$ fraction of magnitude-smallest elements in $\pmb{W}_p^{n+1}$];\\

\end{algorithm}

\subsection{ADMM-based Connection Pruning}

For the connection pruning, the ADMM problem in Equation (\ref{equ:admm1}) can be re-formulated as
\begin{equation}
\underset{\pmb{W}\in P}{\text{min}} ~L=f(\pmb{W})
\label{equ:admm_prune1}
\end{equation}
where $L$ is the normal STBP loss function in Equation (\ref{equ:stbp_loss}) and $P$ denotes a sparse connection space. To match the indicator item in Equation (\ref{equ:admm1}), two steps are required. First, an extra indicator function $g(\pmb{W})$ is added as follows
\begin{equation}
\underset{\pmb{W}}{\text{min}} ~f(\pmb{W})+g(\pmb{W}),
\label{equ:admm_prune2}
\end{equation}
where $g(\pmb{W})=0$ if $\pmb{W}\in P$, $g(\pmb{W})=+\infty$ otherwise. Second, it is further converted to 
\begin{equation}
\underset{\pmb{W},~\pmb{Z}}{\text{min}} ~L=f(\pmb{W})+g(\pmb{Z}),~s.t.~\pmb{W}=\pmb{Z}.
\label{equ:admm_prune3}
\end{equation}
Now the pruning problem is equivalent to the classic ADMM problem given in Equation (\ref{equ:admm1}).

With the constraint of $\pmb{W}=\pmb{Z}$, the augmented Lagrangian can be equivalently simplified to $L_\rho=f(\pmb{W})+g(\pmb{Z})+\pmb{Y}^T(\pmb{W}-\pmb{Z})+\frac{\rho}{2}\|\pmb{W}-\pmb{Z}\|_2^2=f(\pmb{W})+g(\pmb{Z})+\frac{\rho}{2}\|\pmb{W}-\pmb{Z}+\widetilde{\pmb{Y}}\|_2^2-\frac{\rho}{2}\|\widetilde{\pmb{Y}}\|^2_2$ where $\widetilde{\pmb{Y}}=\pmb{Y}/\rho$. In this way, the greedy minimization in Equation (\ref{equ:admm2}) can be re-written as
\begin{equation}
\begin{cases}
\pmb{W}^{n+1}=\underset{\pmb{W}}{\text{arg min}}~L_\rho(\pmb{W},~\pmb{Z}^n,~\widetilde{\pmb{Y}}^n)\\
\pmb{Z}^{n+1}=\underset{\pmb{Z}}{\text{arg min}}~L_\rho(\pmb{W}^{n+1},~\pmb{Z},~\widetilde{\pmb{Y}}^n)\\
\widetilde{\pmb{Y}}^{n+1}=\widetilde{\pmb{Y}}^n+\pmb{W}^{n+1}-\pmb{Z}^{n+1}
\end{cases}.
\label{equ:admm_prune4}
\end{equation}
Actually, the first sub-problem is $\pmb{W}^{n+1}=\underset{\pmb{W}}{\text{arg min}}~f(\pmb{W})+\frac{\rho}{2}\|\pmb{W}-\pmb{Z}^{n}+\widetilde{\pmb{Y}}^n\|_2^2$ which is differentiable and can be directly solved by gradient descent. The second sub-problem is $\pmb{Z}^{n+1}=\underset{\pmb{Z}}{\text{arg min}}~g(\pmb{Z})+\frac{\rho}{2}\|\pmb{W}^{n+1}-\pmb{Z}+\widetilde{\pmb{Y}}^n\|_2^2$, which is equivalent to
\begin{equation}
\underset{\pmb{Z}\in P}{\text{arg min}}~\frac{\rho}{2}\|\pmb{W}^{n+1}-\pmb{Z}+\widetilde{\pmb{Y}}^n\|_2^2.
\label{equ:admm_prune5}
\end{equation}
The above sub-problem can be heuristically solved by keeping a fraction of elements in $(\pmb{W}^{n+1}+\widetilde{\pmb{Y}}^n)$ with the largest magnitudes and setting the rest to zero \cite{zhang2018systematic, zhang2018structadmm, ye2018progressive, ren2019admm}. Given $\pmb{W}^{n+1}$ and $\pmb{Z}^{n+1}$, $\widetilde{\pmb{Y}}$ can be updated according to $\widetilde{\pmb{Y}}^{n+1}=\widetilde{\pmb{Y}}^n+\pmb{W}^{n+1}-\pmb{Z}^{n+1}$. The overall training for ADMM-based connection pruning is provided in Algorithm \ref{alg:admm_prune}. Note that the sparsification step when updating $\pmb{Z}$ is layer-wise rather than network-wise.

\begin{algorithm}
\label{alg:admm_quan}
\caption{ADMM-based Weight Quantization}

\KwData{$b$ (weight bitwidth), $I$ (\#quantization iterations)}

\begin{center}\textbf{Step I: ADMM Retraining for Quantization}\\ \end{center}
Initialize $\pmb{W}^0$ with the pre-trained weights;\\
Initialize $\widetilde{\pmb{Y}}^0=\pmb{0}$;\\
Initialize $\pmb{Z}_q^0=Quan(\pmb{W}^0,~b,~I)$;\\

\KwData{$\pmb{W}^n$, $\pmb{Z}_q^n$, and $\widetilde{\pmb{Y}}^n$ after the $n$-th iteration}
\KwResult{$\pmb{W}^{n+1}$, $\pmb{Z}_q^{n+1}$, and $\widetilde{\pmb{Y}}^{n+1}$}

1. Rewrite the loss function:\\
~~~$L\Leftarrow f(\pmb{W})+\frac{\rho}{2}\|\pmb{W}-\pmb{Z}_q^{n}+\widetilde{\pmb{Y}}^n\|_2^2$;\\
2. Update weights:\\
~~~$\pmb{W}^{n+1}\Leftarrow$ retrain the SNN model one more iteration;\\
3. Update $\pmb{Z}_q^{n+1}$:\\
~~~$\pmb{Z}_q^{n+1}\Leftarrow Quan(\pmb{W}^{n+1}+\widetilde{\pmb{Y}}^n,~b,~I)$;\\
4. Update $\widetilde{\pmb{Y}}^{n+1}$:\\
~~~$\widetilde{\pmb{Y}}^{n+1}\Leftarrow (\widetilde{\pmb{Y}}^n+\pmb{W}^{n+1}-\pmb{Z}_q^{n+1})$;\\

~\\
\begin{center}\textbf{Step II: Hard-Quantization Retraining}\\ \end{center}
Initialize $\pmb{W}_q^0$ with the weights from Step I;\\
Initialize the loss function $L=f(\pmb{W}_q)$;\\

\KwData{$\pmb{W}_q^n$ after the $n$-th iteration}
\KwResult{$\pmb{W}_q^{n+1}$}

Update weights:\\
~~~$\pmb{W}_q^{n+1}\Leftarrow$ retrain the SNN model one more iteration;\\
~~~$\pmb{W}_q^{n+1}\Leftarrow Quan(\pmb{W}_q^{n+1},~b,~I)$;\\

\end{algorithm}

\subsection{ADMM-based Weight Quantization}

The overall framework of ADMM-based weight quantization is very similar to the ADMM-based connection pruning. The only difference is that the constraint on weights changes from the sparse one to a quantized one. Hence, Equation (\ref{equ:admm_prune1}) can be re-written as
\begin{equation}
\underset{\pmb{W}\in Q}{\text{min}} ~L=f(\pmb{W}).
\label{equ:admm_quan1}
\end{equation}
$Q$ is a set of discrete levels, e.g. $Q=\alpha \{0,~\pm 2^0,~\pm 2^1,...,~\pm 2^{b-1}\}$, where $b$ is the bitwidth and $\alpha$ is a scaling factor that can be independent between layers.  Notice that given the bitwidth $b$, there are $2b+1$ discrete levels with our definition. In the cases of $b\leq 2$, we need $b+1$ bits to store the $2b+1$ levels; while in the cases of $b>2$, we only need $b$ bits or even fewer. For simplicity, we generally denote the needed number of bits as $b$. Although we follow the definition of $Q$ in \cite{leng2018extremely}, our approach still works under other definitions. 

Similarly, now Equation (\ref{equ:admm_prune5}) should be
\begin{equation} 
\underset{\pmb{Z}\in Q}{\text{arg min}}~\frac{\rho}{2}\|\pmb{W}^{n+1}-\pmb{Z}+\widetilde{\pmb{Y}}^n\|_2^2,
\label{equ:admm_quan2}
\end{equation}
which is equivalent to 
\begin{equation} 
\underset{\widetilde{\pmb{Z}}, \alpha}{\text{arg min}}~\frac{\rho}{2}\|\pmb{V}-\alpha \widetilde{\pmb{Z}}\|_2^2
\label{equ:admm_quan3}
\end{equation}
where $\pmb{V}=\pmb{W}^{n+1}+\widetilde{\pmb{Y}}^n$, $\alpha \widetilde{\pmb{Z}}=\pmb{Z}$, and $\widetilde{\pmb{Z}}\subset \{0,~\pm 2^0,~\pm 2^1,...,~\pm 2^{b-1}\}$. This sub-problem can also be heuristically solved by an iterative quantization \cite{leng2018extremely}, i.e. iteratively fixing $\alpha$ and the quantized vector $\widetilde{\pmb{Z}}$ to convert the bivariate optimization to two iterative univariate optimizations. Specifically, with $\alpha$ fixed, the quantized vector $\widetilde{\pmb{Z}}$ is actually the projection of $\frac{\pmb{V}}{\alpha}$ onto $\{0,~\pm 2^0,~\pm 2^1,...,~\pm 2^{b-1}\}$, which can be simply obtained by approaching the closest discrete level of each element; with $\widetilde{\pmb{Z}}$ fixed, $\alpha$ can be easily calculated by $\alpha=\frac{\pmb{V}^{T}\widetilde{\pmb{Z}}}{{\widetilde{\pmb{Z}}}^{T}\widetilde{\pmb{Z}}}$. In practice, we find this iterative minimization converges very fast (e.g. in three iterations). The overall training for ADMM-based weight quantization is given in Algorithm \ref{alg:admm_quan}, where the quantization function ($Quan(\cdot)$) is additionally given in Algorithm \ref{alg:quan}. Note that the quantization step when updating $\pmb{Z}$ is layer-wise too, which might cause different $\alpha$ values across layers.

\begin{algorithm}
\label{alg:quan}
\caption{Quantization Function - $Quan(\cdot)$}

\KwData{$\pmb{V}$, $b$ (weight bitwidth), $I$ (\#quantization iterations)}
\KwResult{$\pmb{Z}$}

Define a discrete space $Q=\{0,~\pm 2^0,~\pm 2^1,...,~\pm 2^{b-1}\}$;\\
Initialize $\alpha=1$;\\

~\\
\For{$i=0:I-1$}{

1. Update $\widetilde{\pmb{Z}}$:\\
~~~$\widetilde{\pmb{Z}}\Leftarrow$ project each element in $\frac{\pmb{V}}{\alpha}$ to its nearest discrete level in $Q$;\\
2. Update $\alpha$:\\
~~~$\alpha \Leftarrow \frac{\pmb{V}^{T}\widetilde{\pmb{Z}}}{{\widetilde{\pmb{Z}}}^{T}\widetilde{\pmb{Z}}}$;\\

}
$\pmb{Z}\Leftarrow \alpha \widetilde{\pmb{Z}}$;\\

\end{algorithm}

\begin{figure}[!htbp]
\centering
\includegraphics[width=0.3\textwidth]{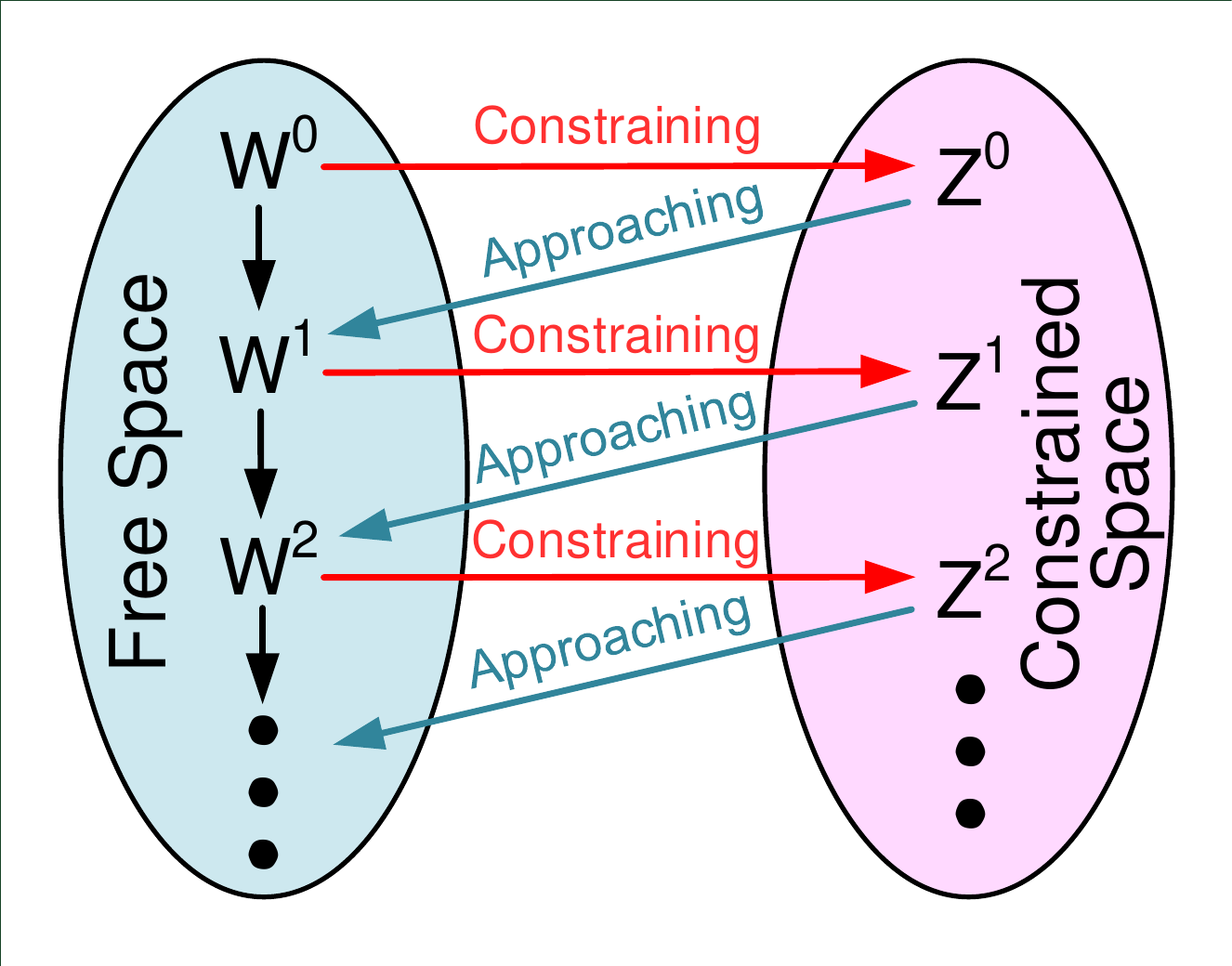}
\caption{Weight space evolution during ADMM training. At each iteration, $\pmb{Z}$ is obtained by constraining $\pmb{W}$ into a constrained space and $\pmb{W}$ gradually approaches $\pmb{Z}$.}
\label{fig:admm_space}
\end{figure}

\begin{figure}[!htbp]
    \centering
    \includegraphics[width=0.495\textwidth]{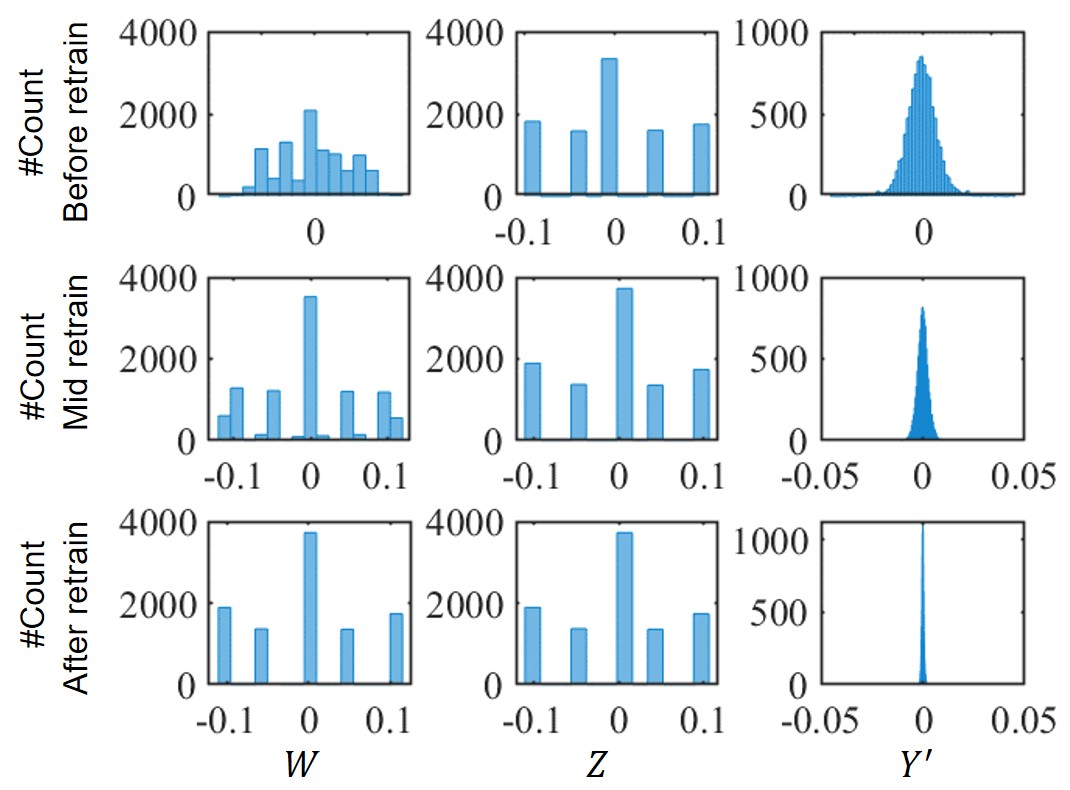}
   \caption{Evolution of $\pmb{W}$, $\pmb{Z}$, $\widetilde{\pmb{Y}}$ during ADMM retraining.}
   \label{fig:admm_evolution}
   \vspace{-10pt}
\end{figure}

Figure \ref{fig:admm_space} presents the evolution of the weight space during ADMM retraining. In fact, $\pmb{Z}$ strictly satisfies the constraints (sparse or quantized) at each iteration by solving Equation (\ref{equ:admm_prune5}) or (\ref{equ:admm_quan2}), respectively. Moreover, $\pmb{W}$ gradually approaches $\pmb{Z}$ by minimizing the $L2$-norm regularizer, i.e. $\frac{\rho}{2}\|\pmb{W}-\pmb{Z}^{n}+\widetilde{\pmb{Y}}^n\|_2^2$, in the first sub-problem of ADMM. The auxiliary variable $\widetilde{\pmb{Y}}$ tends to be zero (we omit it in Figure \ref{fig:admm_space} for simplicity). To evidence the above prediction, we visualize the distributions of $\pmb{W}$, $\pmb{Z}$, and $\widetilde{\pmb{Y}}$ at different stages during the entire ADMM retraining process, as depicted in Figure \ref{fig:admm_evolution}. Here we take the weight quantization as an example and set $\rho=0.1$.  Apparently, $\pmb{Z}$ is always in a quantized space with limited levels while $\pmb{W}$ gradually approaches $\pmb{Z}$ as ADMM retraining goes on. Notice that although the distributions of $\pmb{Z}$ seem similar and unchanged due to the limited weight levels, its value at each element indeed changes until convergence.  Compared to the hard pruning \cite{han2015learning, han2015deep} or quantization \cite{wu2018training, deng2018gxnor}, ADMM-based compression is able to achieve better convergence due to the multivariable optimization.

\subsection{Activity Regularization}

As aforementioned in Section \ref{sec:compression_way}, the compute cost of SNNs is jointly determined by the baseline operations and the average spike rate during runtime. Hence, besides the connection pruning and the weight quantization, there is an extra opportunity to reduce the compute cost by activity regularization. To this end, we tune the loss function to
\begin{equation}
\label{equ:regu_loss}
L = L_{normal} + \lambda R
\end{equation}
where $L_{normal}$ is the vanilla loss function in Equation (\ref{equ:stbp_loss}), $R$ is the mentioned average spike rate per neuron per timestep, and $\lambda$ is a penalty coefficient. The reason that we use the average spike rate rather than the total number of spikes is to unify the exploration of $\lambda$ setting across different networks. By introducing the above regularization item, we can further sparsify the firing activities of an SNN model, resulting in decreased active operations.

In essence, a similar but different work was pioneered in \cite{neil2016learning}. It adopted the mentioned ANN-to-SNN-conversion learning method that is distinct from our direct training of SNNs. The activity regularization in \cite{neil2016learning} was applied on the activations of the ANN model during training, indirectly lowering the number of spikes in the resulting SNN model converted from the ANN model during inference. The effectiveness of the approach is based on an expectation that the neuronal activation of an ANN neuron is proportional to the spike rate of its converted SNN neuron. By contrast, our activity regularization is directly applied on the spikes of neurons in the SNN model, which does not rely on such an assumption.

\subsection{Compression Strategy: Single-way or Joint-way}

Based on the ADMM-based connection pruning and weight quantization, as well as the activity regularization, we propose two categories of compression strategy: single-way and joint-way. Specifically, i) \textbf{single-way compression} individually applies connection pruning, weight quantization, or activity regularization; ii) \textbf{joint-way compression} jointly applies connection pruning, weight quantization, and activity regularization, including ``Pruning \& Regularization'', ``Quantization \& Regularization'', ``Pruning \& Quantization'', and ``Pruning \& Quantization \& Regularization''. Compared to the single-way compression, the joint-way compression can usually achieve a more aggressive overall compression ratio by exploiting multiple information ways. 

For ``Pruning \& Regularization'' and ``Quantization \& Regularization'', we introduce the activity regularization item $\lambda R$ into the loss functions in Algorithm \ref{alg:admm_prune} and Algorithm \ref{alg:admm_quan}, respectively. For ``Pruning \& Quantization'', we merge both the connection pruning and the weight quantization, as presented in Algorithm \ref{alg:admm_prune_quan}. For ``Pruning \& Quantization \& Regularization'', we further incorporate the activity regularization item into the loss function in Algorithm \ref{alg:admm_prune_quan}.

\begin{algorithm}
\label{alg:admm_prune_quan}
 \caption{ADMM-based Pruning \& Quantization }

\KwData{$s$ (connection sparsity), $b$ (weight bitwidth), $I$ (\#quantization iterations)}

\begin{center}\textbf{Step I: ADMM Retraining for Pruning}\\ \end{center}
Initialize $\pmb{W}^0$ with the pre-trained weights;\\
Generate sparse weights $\pmb{W}_p$ by retraining the SNN model with Algorithm \ref{alg:admm_prune};\\
Generate a binary mask $\pmb{M}_p$ in which 1s and 0s denote the remained and pruned weights in $\pmb{W}_p$, respectively;\\

\begin{center}\textbf{Step II: ADMM Retraining for Pruning \& Quantization}\\ \end{center}
Initialize $\pmb{W}^0$ with the weights from Step I;\\
Initialize $\widetilde{\pmb{Y}}^0=\pmb{0}$;\\
Initialize $\pmb{Z}_{pq}^0=Quan(\pmb{W}^0,~b,~I)$;\\

\KwData{$\pmb{W}^n$, $\pmb{Z}_{pq}^n$, and $\widetilde{\pmb{Y}}^n$ after the $n$-th iteration}
\KwResult{$\pmb{W}^{n+1}$, $\pmb{Z}_{pq}^{n+1}$, and $\widetilde{\pmb{Y}}^{n+1}$}

1. Rewrite the loss function:\\
~~~$L\Leftarrow f(\pmb{W})+\frac{\rho}{2}\|\pmb{W}-\pmb{Z}_{pq}^{n}+\widetilde{\pmb{Y}}^n\|_2^2$;\\
2. Update weights:\\
~~~$\pmb{W}^{n+1}\Leftarrow$ retrain the SNN model one more iteration (update only the non-zero weights according to $\pmb{M}_p$);\\
3. Update $\pmb{Z}_{pq}^{n+1}$:\\
~~~$\pmb{Z}_{pq}^{n+1}\Leftarrow Quan(\pmb{W}^{n+1}+\widetilde{\pmb{Y}}^n,~b,~I)$;\\

4. Update $\widetilde{\pmb{Y}}^{n+1}$:\\
~~~$\widetilde{\pmb{Y}}^{n+1}\Leftarrow (\widetilde{\pmb{Y}}^n+\pmb{W}^{n+1}-\pmb{Z}_{pq}^{n+1})$;\\

\begin{center}\textbf{Step III: Hard-Pruning-Quantization Retraining}\\ \end{center}
Initialize $\pmb{W}_{pq}^0$ with the weights from Step II;\\
Initialize the loss function $L=f(\pmb{W}_{pq})$;\\

\KwData{$\pmb{W}_{pq}^n$ after the $n$-th iteration}
\KwResult{$\pmb{W}_{pq}^{n+1}$}

Update weights:\\
~~~$\pmb{W}_{pq}^{n+1}\Leftarrow$ retrain the SNN model one more iteration (update only the non-zero weights according to $\pmb{M}_p$);\\
~~~$\pmb{W}_{pq}^{n+1}\Leftarrow Quan(\pmb{W}_{pq}^{n+1},~b,~I)$;\\

\end{algorithm}

\subsection{Quantitative Evaluation Metrics for SNN Compression}\label{sec:metric}

The compression ratio can be reflected by the reduced memory and compute costs. On the memory side, we just count the required storage space for weight parameters since they occupy the most memory space. On the compute side, we just count the required addition operations because multiplications can be removed from SNNs with binary spike representation. The connection pruning reduces the number of parameters and baseline operations, thus lowering both the memory and compute costs; the weight quantization reduces the bitwidth of parameters and the basic cost of each addition operation, thus also lowering both the memory and compute costs; the activity regularization reduces the number of dynamic spikes, thus mainly lowering the compute cost.  Note that although fewer spikes can also reduce the associated memory access, the memory traffic cost depends on actual hardware architectures, which is out the scope of this algorithm-level paper. Therefore, we only discuss the reduction of the compute cost when applying activity regularization. 

Here we propose several metrics to quantitatively evaluate the compression ratio of SNNs. For the memory compression, we define the following percentage of residual memory cost:
\begin{equation}
\label{equ:mem_compression}
R_{mem}=(1-s)\cdot b/B
\end{equation}
where $s\in [0,~1]$ is the connection sparsity, $B$ and $b$ are the weight bitwidth of the original model and the compressed model, respectively. Since the operation compression is related to the dynamic spikes, next, we define the percentage of residual spikes as
\begin{equation}
\label{equ:spike_compression}
R_s=r/R
\end{equation}
if we can reduce the average spike rate from $R$ to $r$. Based on the mentioned rule in Section \ref{sec:compression_way} that the total number of operations in an SNN model is calculated by multiplying the number of baseline operations and the average spike rate, we define the percentage of residual operation cost as
\begin{equation}
\label{equ:ops_compression}
R_{ops}\approx (1-s)\cdot \frac{b}{B}\cdot \frac{r}{R}=R_{mem}\cdot R_s.
\end{equation}
Note that above equation is just a coarse estimation because the impact of bitwidth on the operation cost is not linear. For example, an FP32 (i.e. 32-bit floating point) dendrite integration is not strictly 4$\times$ costly than an INT8 (i.e. 8-bit integer) one. 

\section{Experimental Results}\label{sec:results}

\subsection{Experimental Setup}
We validate our compression methodology on various datasets, including the static image datasets (e.g. MNIST, CIFAR10, and \todob{CIFAR100}) and the event-drive N-MNIST, and then observe the compression effect on accuracy and summarize the extent to which an SNN model can be compressed with acceptable functionality degradation. For MNIST and N-MNIST, we use the classic LeNet-5 structure; while for CIFAR10 and  CIFAR100 , we use a convolution with stride of 2 to replace the pooling operation and then design a 10-layer spiking convolutional neural network (CNN) with the structure of Input-128C3S1-256C3S2-256C3S1-512C3S2-512C3S1-1024C3S1-2048C3S2-1024FC-512FC-10/100. We take the Bernoulli sampling to convert the raw pixel intensity to a spike train on MNIST; while on CIFAR10 and  CIFAR100 , inspired by \cite{wu2019direct}, we use an encoding layer to convert the normalized image input into spike trains to improve the baseline accuracy.  The number of presented timesteps (i.e., the number of $dt$) for each input sample is $T$, which is also the time window for the calculation of the average spike rate. The simulation time interval between consecutive sample presentations is assumed to be long enough to let the membrane potential leak sufficiently, avoiding the possible cross-sample interference. The programming environment for our experiments is Pytorch. We omit ``INT'' and only remain the bitwdith for simplicity in the results with weight quantizaiton.

\begin{table}[!htbp]
\centering
\caption{Hyper-parameter setting on different datasets.}
\vspace{2pt}
\label{tab:Config}
\renewcommand\arraystretch{1.3}
\resizebox{0.485\textwidth}{!}{
\begin{tabular}{cccccc}
\hline
Parameters & Descriptions           & MNIST & N-MNIST & CIFAR10 & CIFAR100 \\ \hline
$dt$  &  Duration of Simulation Timetep    & $1ms$   & $1ms$     & $1ms$  &   $1ms$ \\
$N_0$  &  Epochs for Model Pretraining   & 150   & 150     & 100  & 100   \\
Batch Size & --    & 50    & 50      & 50      \\
\multirow{2}*{$T$}    & Number of Presented Timesteps  & \multirow{2}*{10}    & \multirow{2}*{10}     & \multirow{2}*{8}   & \multirow{2}*{8}    \\
&  for Each Sample  & & & & \\
$u_{th}$   & Firing Threshold  & 0.2   & 0.4     & 0.5 & 0.5     \\
$e^{-\frac{dt}{\tau}}$     &  Membrane Potential Decay Factor   & 0.25   & 0.3      & 0.25  & 0.8    \\
$a$        & Gradient Width of  $H(\cdot)$         & 0.5   & 0.5     & 0.5 & 0.5    \\
$\rho$        & Penalty Coefficient for ADMM         & 5e-4   & 5e-4     & 5e-4 & 1e-4    \\ 
\multirow{2}*{$N_1$}       &  Retraining Epochs for 
             & \multirow{2}*{10}   & \multirow{2}*{10}     & \multirow{2}*{20}  & \multirow{2}*{20}   \\
&  ADMM-Compression  & & & & \\
\multirow{2}*{$N_2$}        &  Retraining Epochs for   & \multirow{2}*{10}   & \multirow{2}*{10}   & \multirow{2}*{15}  & \multirow{2}*{20}  \\
& Hard-Compression(HC)  & & & & \\
\hline
\end{tabular}}
\end{table}

First, connection pruning, weight quantization, and activity regularization are evaluated individually for a preliminary effectiveness estimation. Next, the compression is carried out in a joint way with two or three methods, to explore the comprehensive effect and the optimal combination. We do not compress the first and last layers due to their accuracy sensitivity and the insignificant amount of parameters and operations, so the calculation of $R_{mem}$ and $R_{ops}$ does not include these two layers. Details about the hyper-parameter setting can be seen in Table \ref{tab:Config}. 

\subsection{Single-way Compression}

\begin{figure}[!htbp]
\centering
\includegraphics[width=0.48\textwidth]{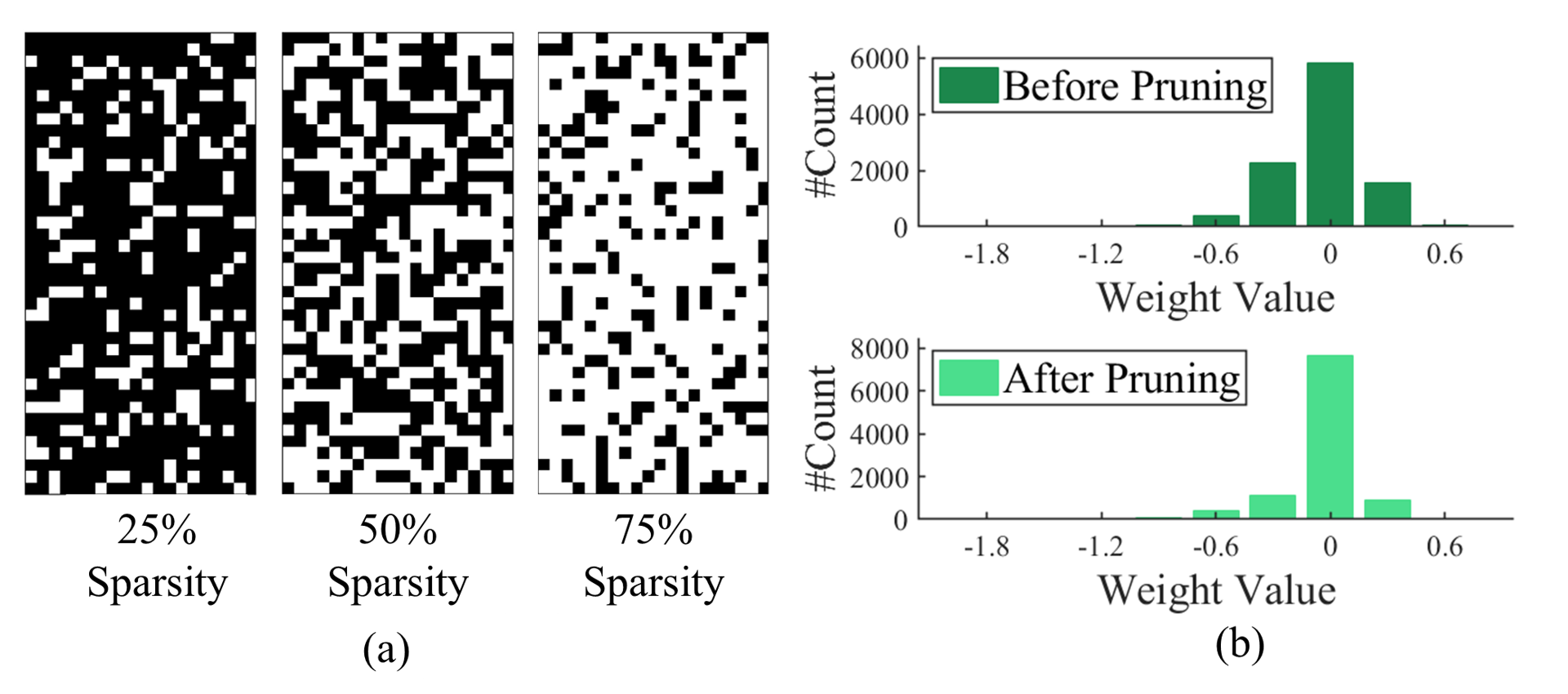}
\caption{Effect of connection pruning on LeNet-5, (a) visualization of 800 randomly selected connections, where white pixels denote pruned connections; (b) weight value distribution before and after pruning with 75\% sparsity.}
\label{fig:Prune_bar}
\vspace{-10pt}
\end{figure}

\begin{figure}[!htbp]
    \centering
    \includegraphics[width=0.42\textwidth]{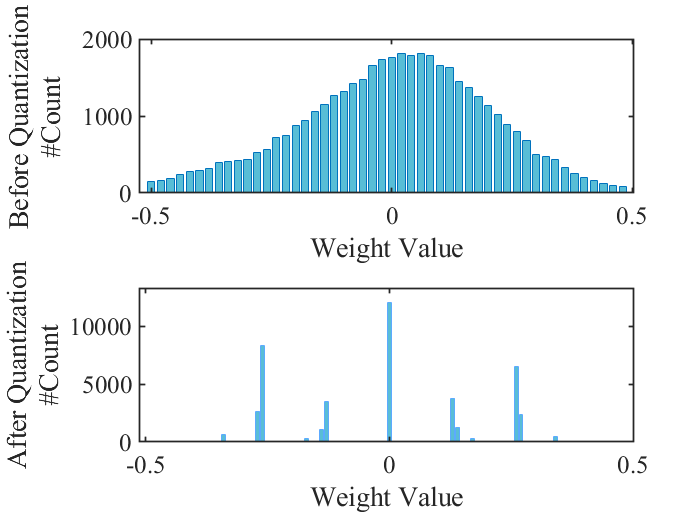}
    \caption{Weight distribution of LeNet-5 before and after weight quantization under $b=2$.}
    \label{fig:Quan_bar}
    \vspace{-10pt}
\end{figure}

\begin{figure}[!htbp]
    \centering
    \includegraphics[width=0.5\textwidth]{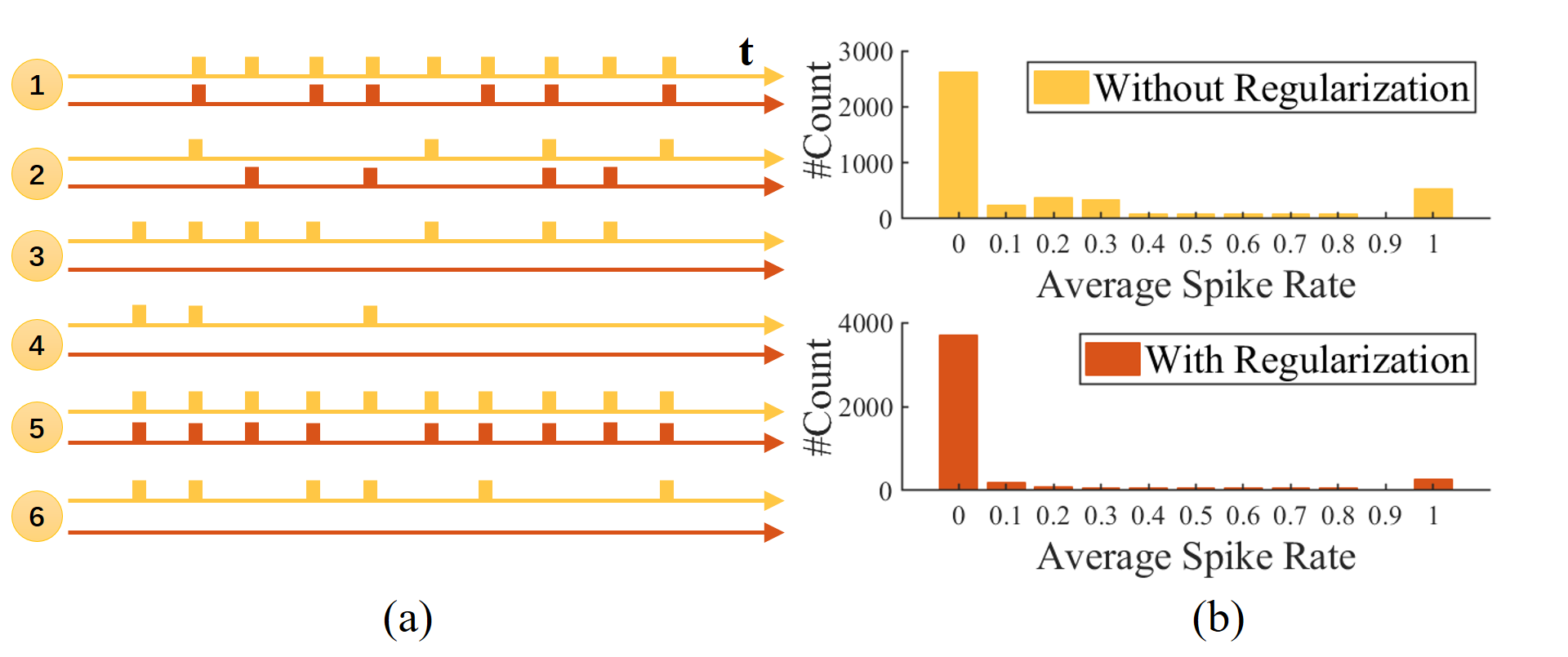}
    \caption{Effect of activity regularization on LeNet-5 ($\lambda=0.01$), (a) spike trains of six randomly selected neurons, where the lower spike train for each neuron is the one after activity regularization; (b) average spike rate distribution without and with activity regularization. Here the average spike rate means the average number of spikes per timestep, and the distribution is across neurons. }
    \label{fig:Regular_bar}
\end{figure}

\begin{figure*}[!htbp]
\centering
\includegraphics[width=0.85\textwidth]{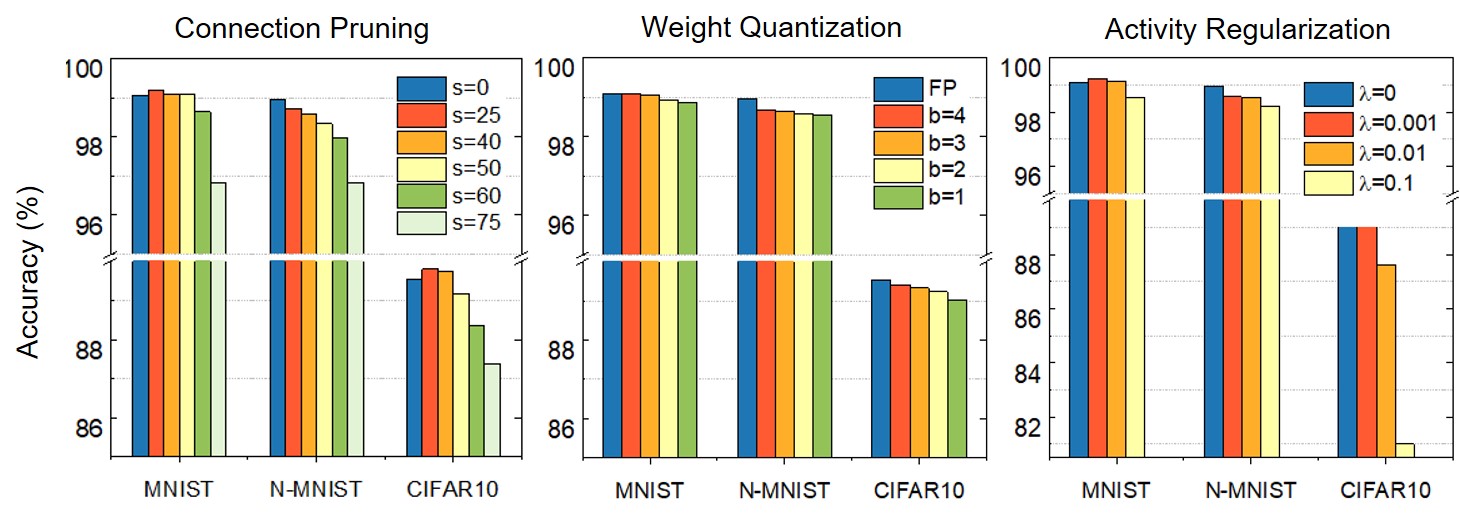}
\caption{ Glance of the accuracy results under single compression.} 
\label{fig:Single_glance}
\end{figure*}

In this subsection, we analyze the results from single-way compression, i.e. applying connection pruning (Figure \ref{fig:Prune_bar} \& Table \ref{tab:single_pruning}), weight quantization (Figure \ref{fig:Quan_bar} \& Table \ref{tab:single_quan}), and activity regularization (Figure \ref{fig:Regular_bar} \& Table \ref{tab:single_act}) individually.  Finally, we present Figure \ref{fig:Single_glance} to summary the accuracy results in Table \ref{tab:single_pruning}-\ref{tab:single_act} for a better glance. 

\begin{table}[!htbp]
\centering
\caption{Accuracy under different connection sparsity.}
\vspace{3pt}
\label{tab:single_pruning}
\renewcommand\arraystretch{1.1}
  \resizebox{0.4\textwidth}{!}{
\begin{tabular}{cccc}
\hline
\hline
Dataset & Sparsity ($s$) & Acc. (\%) & Acc. Loss (\%) \\ 
\hline
\multirow{6}* {MNIST} & 0\% & 99.07 & 0.00 \\
\cline{2-4}
 & 25\% & 99.19 & 0.12 \\
 & 40\% & 99.08 & 0.01 \\
 & 50\% & 99.10 & 0.03 \\
 & 60\% & 98.64 & -0.43 \\
 & 75\% & 96.84 & -2.23 \\
\hline
\multirow{6}* {N-MNIST} & 0\% & 98.95 & 0.00 \\
\cline{2-4}
 & 25\% & 98.72 & -0.23 \\
 & 40\% & 98.59 & -0.36 \\
 & 50\% & 98.34 & -0.61 \\
  & 60\% & 97.98 & -0.91 \\
 & 75\% & 96.83 & -2.12 \\
\hline
\multirow{6}* {CIFAR10} & 0\% & 89.53 & 0.00 \\
\cline{2-4}
 & 25\% & 89.8 & 0.27 \\
 & 40\% & 89.75 &  0.18 \\
 & 50\% & 89.15 & -0.38 \\
 & 60\% & 88.35 & -1.18 \\
 & 75\% & 87.38 & -2.15 \\ 
\hline
\hline
\end{tabular}}
\end{table}

\textbf{Connection Pruning}. As shown in Figure \ref{fig:Prune_bar}, the number of disconnected synapses dramatically increases after connection pruning, and the overall percentage of pruned connections grows accordingly as the sparsity becomes higher. More specifically, Table \ref{tab:single_pruning} shows the model accuracy under different pruning ratio (i.e. connection sparsity). Overall, a $<$40-50\% pruning ratio causes negligible accuracy loss or even better accuracy due to the alleviation of over-fitting, while an over 60\% sparsity would cause obvious accuracy degradation that even reaches  $>$2\% at 75\% sparsity. The accuracy loss on N-MNIST is more severe than that on MNIST, especially in the low-sparsity region. This reflects the accuracy sensitivity to the connection pruning on N-MNIST with natural sparse features. The accuracy on CIFAR10 drops faster due to the increasing difficulty and model size.

\begin{table}[!htbp]
\centering
\caption{Accuracy under different weight bitwidth.}
\vspace{3pt}
\label{tab:single_quan}
\renewcommand\arraystretch{1.1}
  \resizebox{0.4\textwidth}{!}{
\begin{tabular}{ccccc}
\hline
\hline
Dataset & Bitwidth ($b$) & Acc. (\%) & Acc. Loss (\%)\\ 
\hline
\multirow{5}* {MNIST} & 32 (FP) & 99.07 & 0.00 \\
\cline{2-4}
 & 4 & 99.10 & 0.03 \\
 & 3 & 99.04 & -0.03 \\
 & 2 & 98.93 & -0.14 \\
 & 1 & 98.85 & -0.22 \\
\hline
\multirow{5}* {N-MNIST} & 32 (FP) & 98.95 & 0.00 \\
\cline{2-4}
 & 4 & 98.67 & -0.28 \\
 & 3 & 98.65 & -0.30 \\
 & 2 & 98.58 & -0.37 \\
 & 1 & 98.54 & -0.41 \\
\hline
\multirow{5}* {CIFAR10} & 32 (FP) & 89.53 & 0.00 \\ 
\cline{2-4}
 & 4 & 89.40 & -0.13 \\ 
 & 3 & 89.32 & -0.21 \\ 
 & 2 & 89.23 & -0.30 \\
 & 1 & 89.01 & -0.52 \\
\hline
\hline
\end{tabular}}
\end{table}

\textbf{Weight Quantization}. Figure \ref{fig:Quan_bar} evidences that the number of weight levels can be significantly reduced after applying the weight quantization. Note that the number of discrete levels in the network is more than $5$ at $b=2$ due to the different scaling factors (i.e., $\alpha$) across layers after quantization. Moreover, Table \ref{tab:single_quan} presents the accuracy results under different weight bitwidth. On all datasets, we observe negligible accuracy loss when $b\geq 4$. The accuracy loss is still very small ($\leq$0.52\%) even if under the aggressive compression with $b=1$, which reflects the effectiveness of our ADMM-based weight quantization. The accuracy loss on MNIST is constantly smaller than others due to the simplicity of this task.

\begin{table}[!htbp]
\caption{Accuracy under different spike rate.}
\centering
\vspace{3pt}
\label{tab:single_act}
\renewcommand\arraystretch{1.1}
  \resizebox{0.45\textwidth}{!}{
\begin{tabular}{ccccc}
\hline\hline
Dataset & $\lambda$ & Avg. Spike Rate ($r$) & Acc. & Acc. Loss (\%) \\ \hline
\multirow{4}* {MNIST} & 0    &  0.22   &  99.07 & 0.00 \\ \cline{2-5}
 & 0.001 & 0.19 & 99.22  & 0.15\\
 & 0.01 & 0.12 & 99.11   & 0.04\\
 & 0.1 & 0.06 & 98.54    & -0.53\\ \hline
\multirow{4}* {N-MNIST} & 0 & 0.18 & 98.95 &    0.00 \\ \cline{2-5}
 & 0.001 & 0.17 & 98.56 & -0.39\\
 & 0.01 & 0.13 & 98.53  & -0.42\\
 & 0.1 & 0.06 & 98.23  & -0.72\\ \hline
\multirow{4}* {CIFAR10} & 0 & 0.11 & 89.53 & 0.00 \\ \cline{2-5}
 & 0.001 & 0.11 & 89.51 & -0.02\\
 & 0.01 & 0.08 & 87.62  & -1.91\\
 & 0.1 & 0.03 & 81.01 & -8.52\\ \hline\hline
\end{tabular}}
\end{table}

\textbf{Activity Regulaization}. Different from the compression of synapses in previous connection pruning and weight quantization, the activity regularization reduces the number of dynamic spikes thus decreasing the number of active operations. The total number of spike events and the average spike rate can be greatly decreased by using this regularization (see Figure \ref{fig:Regular_bar}). Table \ref{tab:single_act} further lists the accuracy results under different average spike rate, which is realized by adjusting $\lambda$. A larger $\lambda$ leads to a more aggressive regularization, i.e. lower spike rate. From MNIST to N-MNIST and CIFAR10, we observe a gradually weakened robustness to the activity regularization. Also, we find that the baseline average spike rate under $\lambda=0$ gradually decreases, which indicates that a higher baseline spike rate would have more space for activity regularization without compromising much accuracy. This is straightforward to understand because a higher baseline spike rate usually has a stronger capability for initial information representation. 

\begin{table*}[!htbp]
\caption{Accuracy on MNIST when jointly applying two compression methods. }
\centering
\vspace{3pt}
\label{tab:joint_two}
\renewcommand\arraystretch{1.1}
  \resizebox{0.9\textwidth}{!}{
\begin{tabular}{cccccccc}
\hline \hline
$\lambda$ & Sparsity ($s$) & Bitwidth ($b$) & Avg. Spike Rate ($r$) & ${R_{mem}}^{1}$    & ${R_{ops}}^{1}$        & Acc.(\%) & Acc.Loss (\%) \\ \hline
0         & 0\%          & 32 (FP)       & 0.32            & 100.00\% (1.00$\times$)        & 100.00\% (1.00$\times$)        & 99.07    & 0.00          \\ \hline
0.001     & 25\%           & 32 (FP)       & 0.19            & 75.00\% (1.33$\times$)  & 43.14\% (2.32$\times$)  & 99.11    & 0.04          \\
0.001     & 50\%           & 32 (FP)       & 0.18            & 50.00\% (2.00$\times$)  & 28.30\% (3.53$\times$)  & 98.97    & -0.10         \\
0.001     & 75\%           & 32 (FP)       & 0.15            & 25.00\% (4.00$\times$)  & 11.74\% (8.52$\times$)  & 95.30    & -3.77         \\
0.01      & 25\%           & 32 (FP)       & 0.12            & 75.00\% (1.33$\times$)  & 27.52\% (3.63$\times$)  & 99.22    & 0.15          \\
0.01      & 50\%           & 32 (FP)       & 0.12            & 50.00\% (2.00$\times$)  & 18.19\% (5.50$\times$)  & 99.13    & 0.06          \\
0.01      & 75\%           & 32 (FP)        & 0.09            & 25.00\% (4.00$\times$)  & 7.15\% (13.99$\times$)  & 95.39    & -3.68         \\
0.1       & 25\%           & 32 (FP)       & 0.06            & 75.00\% (1.33$\times$)  & 13.52\% (7.40$\times$)  & 98.70    & -0.37         \\
0.1       & 50\%           & 32 (FP)       & 0.06            & 50.00\% (2.00$\times$)  & 8.71\% (11.48$\times$)  & 98.89    & -0.18         \\
0.1       & 75\%           & 32 (FP)       & 0.06            & 25.00\% (4.00$\times$)  & 4.51\% (22.17$\times$)  & 95.49    & -3.66         \\ \hline
0.001     & 0\%            & 3        & 0.21            & 9.38\% (10.66$\times$)  & 5.98\% (16.72$\times$)  & 98.65    & -0.42         \\
0.001     & 0\%            & 2        & 0.22            & 6.25\% (16.00$\times$)  & 4.18\% (23.92$\times$)  & 98.83    & -0.24         \\
0.001     & 0\%            & 1        & 0.20            & 3.13\% (31.95$\times$)  & 1.94\% (51.55$\times$)  & 98.33    & -0.74         \\
0.01      & 0\%            & 3        & 0.13            & 9.38\% (10.66$\times$)  & 3.73\% (26.81$\times$)  & 98.92    & -0.15         \\
0.01      & 0\%            & 2        & 0.12            & 6.25\% (16.00$\times$)  & 2.41\% (41.49$\times$)  & 98.72    & -0.35         \\
0.01      & 0\%            & 1        & 0.12            & 3.13\% (31.95$\times$)  & 1.19\% (84.03$\times$)  & 98.44    & -0.63         \\
0.1       & 0\%            & 3        & 0.06            & 9.38\% (10.66$\times$)  & 1.72\% (58.14$\times$)  & 98.81    & -0.26         \\
0.1       & 0\%            & 2        & 0.07            & 6.25\% (16.00$\times$)  & 1.34\% (74.63$\times$)  & 96.68    & -2.39         \\
0.1       & 0\%            & 1        & 0.05            & 3.13\% (31.95$\times$)  & 0.48\% (208.33$\times$) & 82.25    & -16.82        \\ \hline
0         & 25\%           & 3        & 0.29            & 7.03\% (14.22$\times$)  & 6.32\% (15.82$\times$)  & 98.59    & -0.48         \\
0         & 25\%           & 2        & 0.26            & 4.69\% (21.32$\times$)  & 3.76\% (26.60$\times$)  & 98.75    & -0.32         \\
0         & 25\%           & 1        & 0.22            & 2.34\% (42.74$\times$)  & 1.63\% (61.35$\times$)  & 98.64    & -0.43         \\
0         & 50\%           & 3        & 0.29            & 4.69\% (21.32$\times$)  & 4.26\% (23.47$\times$)  & 98.27    & -0.80         \\
0         & 50\%           & 2        & 0.29            & 3.13\% (31.95$\times$)  & 2.81\% (35.59$\times$)  & 98.53    & -0.54         \\
0         & 50\%           & 1        & 0.22            & 1.56\% (64.10$\times$)  & 1.07\% (93.46$\times$)  & 96.25    & -2.82         \\
0         & 75\%           & 3        & 0.25            & 2.34\% (42.74$\times$)  & 1.84\% (54.35$\times$)  & 97.13    & -1.94         \\
0         & 75\%           & 2        & 0.21            & 1.56\% (64.10$\times$)  & 1.02\% (98.04$\times$)  & 97.01    & -1.96         \\
0         & 75\%           & 1        & 0.18            & 0.78\% (128.21$\times$) & 0.45\% (222.22$\times$) & 93.52    & -5.55         \\ \hline \hline
\end{tabular}}
\begin{tablenotes}
\footnotesize
\item[1] Note$^{1}$: The compression ratio in the parentheses is the reciprocal of $R_{mem}$ or $R_{ops}$.
\end{tablenotes}
\end{table*}

\subsection{Joint-way Compression}

In this subsection, we analyze the results from joint-way compression, i.e. simultaneously applying two or three methods among connection pruning, weight quantization, and activity regularization. Table \ref{tab:joint_two} and \ref{tab:joint_three} provide the accuracy results of the two-way compression and the three-way compression, respectively. Based on these two tables, we summarize several interesting observations as follows.

\begin{figure}[!htbp]
    \centering
    \includegraphics[width=0.48\textwidth]{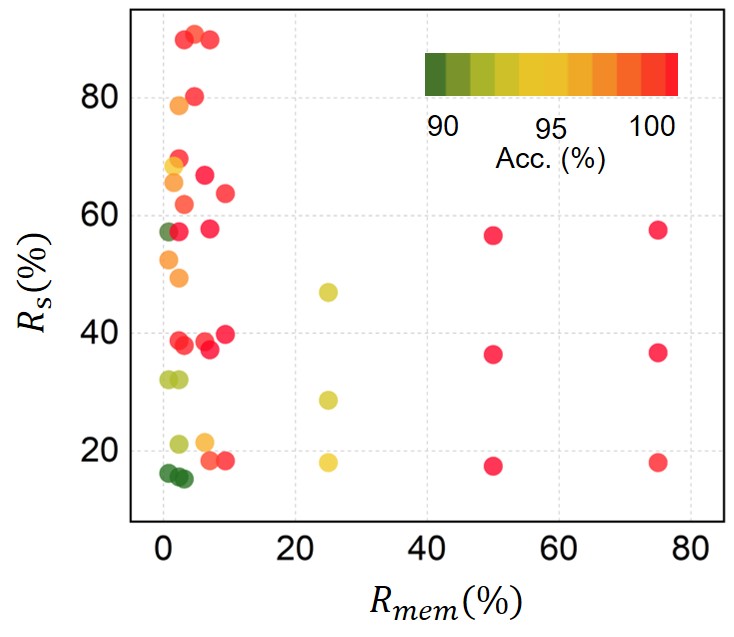}
   \caption{ Accuracy with different $R_{mem}$ and $R_{s}$ in the joint-way compression on MNIST.  All the data are collected from Table \ref{tab:joint_two} and \ref{tab:joint_three}.}
    \label{fig:mem_s}
\end{figure}

\textbf{Contribution to $R_{mem}$ and $R_s$}. The weight quantization contributes most to the reduction of memory (reflected by $R_{mem}$) compared to the connection pruning. For example, an aggressive 75\% connection sparsity (i.e. $R_{mem}$=25\%) just corresponds to a slight 8-bit weight quantization at the same level of memory compression ratio. Note that, as aforementioned, a $b$-bit weight in this work has $2b+1$ discrete levels, which is actually more aggressive quantization than the standard definition with $2^b$ levels when $b>2$. By contrast, the activity regularization contributes most to the reduction of spikes (reflected by $R_s$) because it directly decreases the number of spikes. At last, according to Equation (\ref{equ:ops_compression}), $R_{mem}$ and $R_s$ jointly determine the reduction of operations (reflected by $R_{ops}$).

\begin{table*}[!htbp]
\centering
\caption{ Accuracy on MNIST, CIFAR10 and N-MNIST when applying all three compression methods. }
\vspace{3pt}
\label{tab:joint_three}
\renewcommand\arraystretch{1.1}
  \resizebox{0.95\textwidth}{!}{
\begin{tabular}{ccccccccc}
\hline\hline
Dataset & $\lambda$ & Sparsity ($s$) & Bitwidth ($b$) & Avg. Spike Rate ($r$) & ${R_{mem}}^{1}$   & ${R_{ops}}^{1}$    & Acc. (\%) & Acc. Loss (\%) \\ \hline
\multirow{13}* {MNIST}   & 0         & 0\%              & 32 (FP)             & 0.33                        & 100.00\% (1.00$\times$)        & 100.00\% (1.00$\times$)        & 99.07     & 0.00              \\
\cline{2-9}
   & 0.001     & 25\%               & 3              & 0.19                        & 7.03\% (14.22$\times$)  & 4.06\% (24.63$\times$)  & 98.97     & -0.10           \\
        & 0.001     & 25\%               & 1              & 0.19                        & 2.34\% (42.74$\times$)  & 1.34\% (74.63$\times$)  & 99.04     & -0.03          \\
        & 0.001     & 75\%               & 3              & 0.16                        & 2.34\% (42.74$\times$)  & 1.16\% (86.21$\times$)  & 97.25     & -1.82          \\
        & 0.001     & 75\%               & 1              & 0.17                        & 0.78\% (128.21$\times$) & 0.41\% (243.90$\times$)  & 97.16     & -1.91          \\
   & 0.01      & 25\%               & 3              & 0.12                        & 7.03\% (14.22$\times$)  & 2.61\% (38.31$\times$)  & 99.11     & 0.04           \\
        & 0.01      & 25\%               & 1              & 0.13                        & 2.34\% (42.74$\times$)  & 0.91\% (109.89$\times$) & 98.81     & -0.26          \\
        & 0.01      & 75\%               & 3              & 0.11                        & 2.34\% (42.74$\times$)  & 0.75\% (133.33$\times$) & 94.92     & -4.15          \\
        & 0.01      & 75\%               & 1              & 0.11                        & 0.78\% (128.21$\times$) & 0.25\% (400.00$\times$)    & 94.56     & -4.51          \\
   & 0.1       & 25\%               & 3              & 0.06                        & 7.03\% (14.22$\times$)  & 1.29\% (77.52$\times$)  & 98.25     & -0.82          \\
        & 0.1       & 25\%               & 1              & 0.05                        & 2.34\% (42.74$\times$)  & 0.36\% (277.78$\times$) & 88.14     & -10.93         \\
        & 0.1       & 75\%               & 3              & 0.07                        & 2.34\% (42.74$\times$)  & 0.49\% (204.08$\times$) & 94.54     & -4.43          \\
        & 0.1       & 75\%               & 1              & 0.06                        & 0.78\% (128.21$\times$) & 0.13\% (769.23$\times$) & 74.98     & -24.09         \\ \hline
 
\multirow{13}* { CIFAR10}   & 0\%     & 0    & 32 (FP) & 0.110 & 100.00\% (1.00$\times$) & 100.00\% (1.00$\times$) & 89.53 & 0.00      \\
 \cline{2-9} & 0.001 & 25\%   & 3       & 0.10 & 7.03\% (14.22$\times$)  & 6.14\% (16.29$\times$)  & 87.84 & -1.69  \\
  & 0.001 & 25\%   & 1       & 0.09 & 2.34\% (42.74$\times$)  & 1.81\% (55.25$\times$)  & 87.42 & -2.11  \\
  & 0.001 & 75\%   & 3       & 0.09 & 2.34\% (42.74$\times$)  & 1.85\% (54.05$\times$)  & 87.59 & -1.94  \\
   & 0.001 & 75\%   & 1       & 0.09 & 0.78\% (128.21$\times$) & 0.6\% (166.67$\times$)  & 86.99 & -2.54  \\
 & 0.01  & 25\%   & 3       & 0.06 & 7.03\% (14.22$\times$)  & 3.71\% (26.95$\times$)  & 87.37 & -2.16  \\
 & 0.01  & 25\%   & 1       & 0.06 & 2.34\% (42.74$\times$)  & 1.3\% (76.92$\times$)   & 84.51 & -5.02  \\
& 0.01  & 75\%   & 3       & 0.07 & 2.34\% (42.74$\times$)  & 1.43\% (69.93$\times$)  & 87.13 & -2.40   \\
& 0.01  & 75\%   & 1       & 0.06 & 0.78\% (128.21$\times$) & 0.39\% (256.41$\times$) & 86.75 & -2.78  \\
 & 0.1   & 25\%   & 3       & 0.04 & 7.03\% (14.22$\times$)  & 2.24\% (44.64$\times$)  & 82.75 & -6.78  \\
 & 0.1   & 25\%   & 1       & 0.03 & 2.34\% (42.74$\times$)  & 0.66\% (151.52$\times$) & 77.78 & -11.75 \\
 & 0.1   & 75\%   & 3       & 0.04 & 2.34\% (42.74$\times$)  & 0.74\% (135.14$\times$) & 80.64 & -8.89  \\
 & 0.1   & 75\%   & 1       & 0.03 & 0.78\% (128.21$\times$) & 0.22\% (454.55$\times$) & 74.83 & -14.70 \\ \hline
        
\multirow{13}* {N-MNIST} & 0         & 0\%              & 32 (FP)             & 0.19        & 100.00\% (1.00$\times$)        & 100.00\% (1.00$\times$)         & 98.95     & 0.00              \\
\cline{2-9}
 & 0.001     & 25\%               & 3              & 0.03                        & 7.03\% (14.22$\times$)  & 1.29\% (77.52$\times$)  & 98.62     & -0.33          \\
        & 0.001     & 25\%               & 1              & 0.03                        & 2.34\% (42.74$\times$)  & 0.43\% (232.56$\times$) & 98.57     & -0.38          \\
        & 0.001     & 75\%               & 3              & 0.03                        & 2.34\% (42.74$\times$)  & 0.35\% (285.71$\times$) & 96.19     & -2.76          \\
        & 0.001     & 75\%               & 1              & 0.03                        & 0.78\% (128.21$\times$) & 0.13\% (769.23$\times$) & 96.33     & -2.62          \\
 & 0.01      & 25\%               & 3              & 0.03                        & 7.03\% (14.22$\times$)  & 0.97\% (103.09$\times$) & 98.73     & -0.22          \\
        & 0.01      & 25\%               & 1              & 0.03                        & 2.34\% (42.74$\times$)  & 0.32\% (312.50$\times$)  & 98.66     & -0.29          \\
        & 0.01      & 75\%               & 3              & 0.02                        & 2.34\% (42.74$\times$)  & 0.28\% (357.14$\times$) & 97.23     & -1.72          \\
        & 0.01      & 75\%               & 1              & 0.02                        & 0.78\% (128.21$\times$) & 0.1\% (1000.00$\times$)    & 97.19     & -1.76          \\
 & 0.1       & 25\%               & 3              & 0.01                        & 7.03\% (14.22$\times$)  & 0.42\% (238.10$\times$)  & 98.43     & -0.52          \\
        & 0.1       & 25\%               & 1              & 0.01                        & 2.34\% (42.74$\times$)  & 0.14\% (714.29$\times$) & 98.37     & -0.58          \\
        & 0.1       & 75\%               & 3              & 0.01                        & 2.34\% (42.74$\times$)  & 0.12\% (833.33$\times$) & 96.74     & -2.21          \\
        & 0.1       & 75\%               & 1              & 0.01                        & 0.78\% (128.21$\times$) & 0.04\% (2500.00$\times$)   & 96.87     & -2.08          \\ \hline\hline
\end{tabular}
}
\begin{tablenotes}
\footnotesize
\item[1] Note$^{1}$: The compression ratio in the parentheses is the reciprocal of $R_{mem}$ or $R_{ops}$.
\end{tablenotes}
\end{table*}

\begin{figure}[!htbp]
    \centering
    \includegraphics[width=0.45\textwidth]{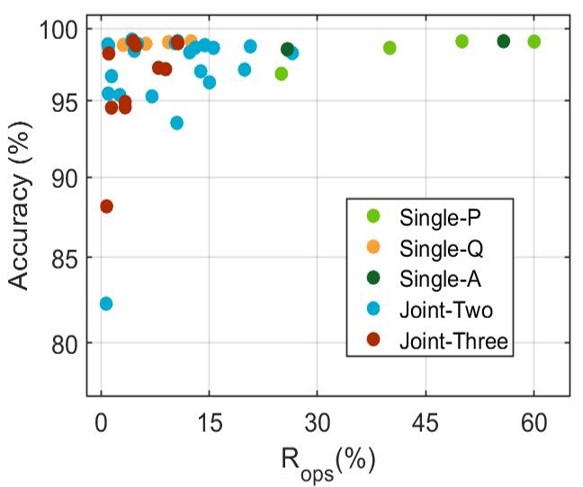}
   \caption{Relationship between $R_{ops}$ and accuracy on MNIST.  All data are collected from Table \ref{tab:single_pruning}-\ref{tab:joint_three}. Abbreviations: P-connection pruning, Q-weight quantization, A-activity regularization,  ``Joint-Two''--joint-way compression with two single-way compression methods (i.e., pruning \& regularization, quantization \& regularization, and pruning \& quantization), ``Joint-Three''--joint-way compression with all three single-way compression methods (i.e., pruning \& quantization \& regularization).} 
   \label{fig:ops_acc}
\end{figure}

\textbf{Trade-off between $R_{mem}$ and $R_s$}. The compression ratios of synapse memory and dynamic spikes actually behave as a trade-off. A too aggressive spike compression baseline (e.g. under $\lambda=0.1$) will cause a large accuracy loss when $R_{mem}$ slightly decreases; a too aggressive memory compression baseline (e.g. $R_{mem}<$5\%) will also cause a significant accuracy loss when $R_s$ slightly decreases. It is challenging to aggressively compress both the synapse memory and dynamic spikes without compromising accuracy. Figure \ref{fig:mem_s} evidences this trade-off by visualizing all the joint-way compression results from Table \ref{tab:joint_two} and \ref{tab:joint_three} on the $R_{mem}$-$R_s$ plane.  Notice that the point ($R_{mem}=0.78$, $R_s=56.25\%$ without activity regularization achieves lower accuracy (93.52\%) than its neighboring points with higher $R_{mem}$ values or with the same $R_{mem}$ value but using a slight activity regularization (e.g., $\lambda=0.001$, 97.16\% and $\lambda=0.01$, 94.56\%). This reflects that a slight activity regularization can improve accuracy, somehow like the dropout technique.  Furthermore, since we have the $R_{ops}$ metric that takes both $R_{mem}$ and $R_s$ into account (see Equation (\ref{equ:ops_compression})), we further visualize the relationship between $R_{ops}$ and accuracy, which is depicted in Figure \ref{fig:ops_acc}.  Note that the data here are collected from  Table \ref{tab:single_pruning}-\ref{tab:joint_three}. It can be seen that, from a global angle, accuracy is positively correlated to $R_{ops}$, i.e. a lower $R_{ops}$ is prone to cause a lower accuracy. However, the local relationship between $R_{ops}$ and accuracy loses monotonicity to some extent and shows slight variance. The underlying reason is due to the imbalanced influences on accuracy of different single-way compression methods. Even if keeping the $R_{ops}$ values very close, it is possible to get variable accuracy scores by adopting different strategies to combine the single-way compression methods, let alone in the cases of obviously different $R_{ops}$ values in Figure \ref{fig:ops_acc}. The imbalance is even increased by the weight quantization that attracts accuracy towards the top-left direction, thus increasing the accuracy variance when we look at all the joint-compression points. 

\textbf{Joint-way Compression v.s. Single-way Compression}. We recommend to gently compress multi-way information rather than to aggressively compress only single-way information. Specifically, an aggressive compression in one way (e.g. $\geq$75\% connection sparsity, $1$-bit weight bitwidth, or $\lambda\geq 0.1$ activity regularization) is easy to cause accuracy collapse. In contrast, a gentle compression in each of the multiple ways is able to produce a better overall compression ratio while paying smaller accuracy loss. For example, the accuracy loss is only $0.26$\% on MNIST when concurrently applying 25\% connection pruning, $1$-bit weight quantization, and $\lambda=0.01$ activity regularization. In this case, the overall compression ratio actually reaches as aggressive as $R_{mem}=2.34$\% (i.e. 42.74$\times$ compression) and $R_{ops}=0.91$\% (i.e. 109.89$\times$ compression). If we expect the same $R_{ops}$ using the single-way compression, the accuracy would drop dramatically. Figure \ref{fig:ops_acc} reflects this guidance too, where the joint-way compression can reduce more $R_{ops}$ with the same level of accuracy loss.

\begin{figure}[!htbp]
    \centering
    \includegraphics[width=0.48\textwidth]{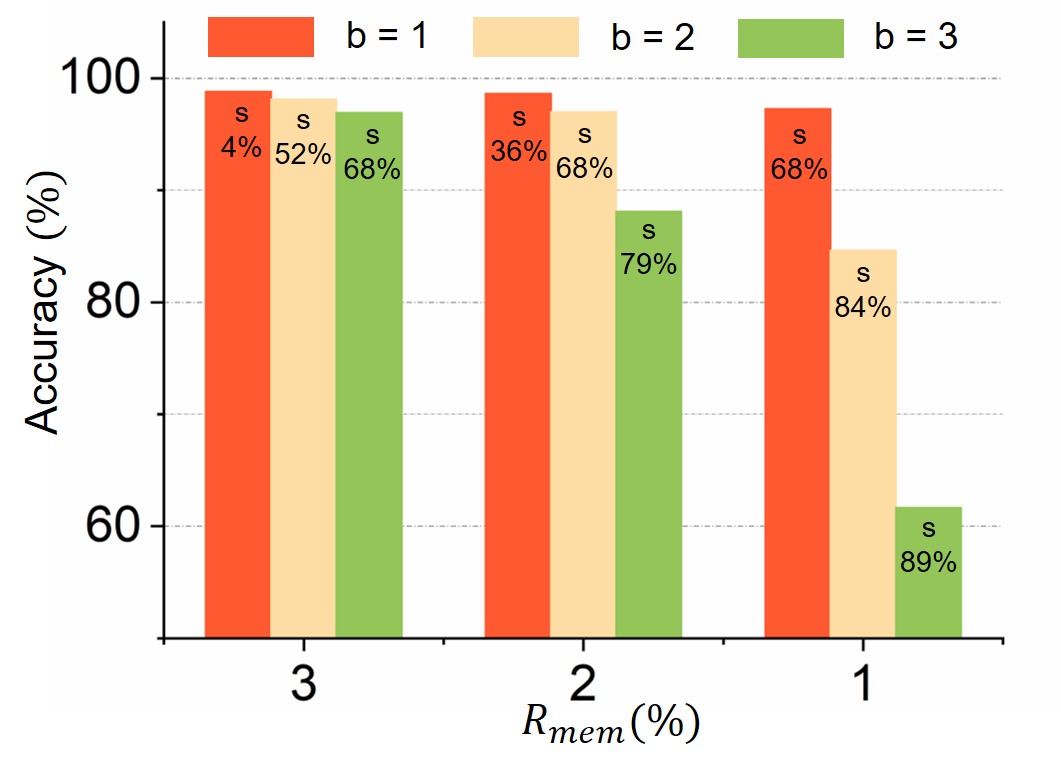}
    \caption{ Accuracy on MNIST under different weight compression ratio and strategy. $R_{mem}$ is controlled by connection sparsity ($s$) and weight bitwidth ($b$) according to Equation (\ref{equ:mem_compression}).} 
    \label{fig:Prune_quan}
    \vspace{-10pt}
\end{figure}

\textbf{Accuracy Tolerance to Weight Quantization}. Recalling Table \ref{tab:single_pruning} and \ref{tab:single_quan}, we observe that SNN models usually present a better accuracy tolerance to the weight quantization than the connection pruning. For example, the accuracy loss at 75\% connection sparsity could reach $>$2\%, while the loss at 1-bit weight quantization is only $<$0.55\%. We use Figure \ref{fig:Prune_quan} to further evidence this speculation. It can be seen that, under the same weight compression ratio, we find the ``aggressive quantization \& slight pruning'' schemes are able to maintain accuracy better than the ``slight quantization \& aggressive pruning'' schemes.

\textbf{Robustness on N-MNIST Dataset}. From Table \ref{tab:joint_three}, we find that SNNs on N-MNIST present more graceful accuracy loss against the joint compression than those on other datasets we used, especially in the cases of aggressive compression. For instance, the accuracy loss is only about 2\% even if an extremely aggressive compression of $R_{mem}=0.78$\% (i.e. 128.21$\times$ compression) and $R_{ops}=0.04$\% (i.e. 2500$\times$ compression) is applied. By contrast, this degree of compression on MNIST will cause $>$20\% accuracy loss. Recalling our observations in single-way compression, SNNs on N-MNIST are more prone to cause higher accuracy degradation than the ones on MNIST under low compression ratios. Considering these together, we expect that the underlying reason lies in the sparse features within these event-driven datasets (e.g. N-MNIST), where the information is heavily scattered along the temporal dimension. This temporal scattering of information causes accuracy degradation when the model meets any compression (even if the compression is slight) due to the sensitivity of intrinsic sparse features, while significantly reducing the accuracy drop when facing an aggressive compression owing to the lower data requirement to represent sparse features.

\subsection{Effectiveness of ADMM optimization}

In this subsection, we verify the effectiveness of ADMM optimization. We compare the accuracy results between the cases with and without ADMM optimization. Note that the compression without ADMM optimization is named hard compression (HC) here. The results are presented in Table \ref{tab:hard_compression} that involves both single-way and joint-way compression. The accuracy scores with ADMM optimization are consistently better than those without ADMM optimization, which evidences the effectiveness of our ADMM optimization in SNN compression.

\begin{table}[!htbp]
\caption{  Comparison to hard compression (HC) without ADMM optimization on CIFAR10. }
\vspace{3pt}
\centering
\label{tab:hard_compression}
\renewcommand\arraystretch{1.1}
  \resizebox{0.485\textwidth}{!}{
\begin{tabular}{ccccc}
\hline\hline
Sparsity ($s$) & Bitwidth ($b$) & $R_{mem}$  & Acc. of HC (\%) & Acc. of Ours (\%) \\\hline
50\%          & 32 (FP)       & 50.00\% & 88.98            & 89.15      \\
75\%          & 32 (FP)       & 25.00\% & 87.00             & 87.38      \\ 
90\%          & 32 (FP)       & 10.00\% & 77.90             & 85.68            \\\hline
0\%           & 3            & 9.38\%  & 88.98            & 89.32      \\
0\%           & 2            & 6.25\%  & 88.78            & 89.23      \\
0\%           & 1            & 3.13\%  & 88.63            & 89.01      \\\hline 
75\%          & 3            & 2.34\%  & 80.30             & 87.27      \\
90\%          & 3            & 0.94\%  & 75.47            & 83.35       \\  
75\%          & 1            & 0.78\%  & 78.85            & 86.71      \\
\hline\hline
\end{tabular}} 
\end{table}

\subsection{Comparison with Existing SNN Compression}

Table \ref{tab:comparison_snn} compares our results with other existing works that touch SNN compression. A fair comparison should take both the compression ratio and recognition accuracy into account. In this sense, our approach is able to achieve a much higher overall compression performance, owing to the accurate STBP learning and the powerful ADMM optimization. Note two points: (1) the recent ReStoCNet \cite{srinivasan2019restocnet} is not a pure SNN model where the FC layers use non-spiking neurons and do not apply any compression technique, which significantly contributes to accuracy maintaining; (2) the models in \cite{Bodo2017Conversion} are trained using the ANN-to-SNN-conversion approach, where the required $T$ is usually much larger than ours.

\begin{table}[!htbp]
\caption{ Comparison with existing SNN compression works on MNIST, CIFAR10, or  CIFAR100  (the default dataset is MNIST unless otherwise specified). Here the number of layers includes the input layer.}
\vspace{3pt}
\label{tab:comparison_snn}
\renewcommand\arraystretch{1.1}
  \resizebox{0.485\textwidth}{!}{
\begin{tabular}{ccccc}
\hline\hline
  & Net. Structure & Sparsity ($s$) & Bitwidth ($b$) & Acc. (\%) \\ \hline
Spiking DBN \cite{Stromatias2015Robustness} & 4-layer MLP & 0\% & 4 & 91.35   \\
Pruning \& Quantization \cite{rathi2018stdp} & 2-layer  MLP$^{1}$ & 92\% & ternary & 91.50  \\
Soft-Pruning \cite{shi2019soft} & 3-layer  MLP & 75\% & 32 (FP) & 94.05  \\
Stochastic-STDP \cite{yousefzadeh2018practical} & 3-layer  MLP & 0\% & binary$^{2}$ & 95.70   \\
NormAD \cite{Kulkarni2018Spiking} & 3-layer  CNN & 0\% & 3 & 97.31   \\
ReStoCNet \cite{srinivasan2019restocnet} & 5-layer CNN$^{3}$ & 0\% & binary$^{4}$ & 98.54 \\
ReStoCNet (CIFAR10) \cite{srinivasan2019restocnet} & 5-layer CNN$^{3}$ & 0\% & binary$^{4}$ & 66.23 \\
 Spiking CNN (CIFAR10) \cite{Bodo2017Conversion} $^{5}$ & 9-layer CNN & 0\% & binary & 83.35/87.45$^{6}$ \\
Spiking CNN (CIFAR100) \cite{esser2016convolutional} & N.A. &  0\% & ternary & 55.64 \\
This work & 5-layer  CNN & 0\% & 3 & 99.04  \\ 
This work & 5-layer  CNN & 25\% & 1 & 98.81$^{7}$  \\ 
This work (CIFAR10) & 5-layer  CNN$^{8}$ & 50\% & 1 & 68.52  \\ 
This work (CIFAR10) & 11-layer  CNN  & 50\% & 1 & 87.21  \\ 
This work (CIFAR100) & 11-layer  CNN  & 50\% & 3 & 57.83  \\ 
This work (CIFAR100) & 11-layer  CNN  & 25\% & 1 & 55.95  \\ 
\hline\hline
\end{tabular}}
\begin{tablenotes}
\footnotesize
\item[1] Note$^{1}$: There is an extra inhibitory layer without compression.
\item[2] Note$^{2}$: The last layer uses 24-bit weight precision.
\item[3] Note$^{3}$: The FC layers use non-spiking neurons.
\item[4] Note$^{4}$: The weights in FC layers are in full precision.
\item[5] Note$^{5}$: The models are trained using the ANN-to-SNN-conversion approach, where the required $T$ is usually much larger than ours.
\item[6] Note$^{6}$: The activations of the ANN before conversion are also binarized.
\item[7] Note$^{7}$: Additional spike compression ($\lambda=0.01$) is applied.
\item[8] Note$^{8}$: For fair comparison, we use the same network structure as \cite{srinivasan2019restocnet} and compress only the Conv layers too. Differently, the neurons in our network are all spiking neurons.
\end{tablenotes}
\vspace{-10pt}
\end{table}
 
\subsection{Comparison with ANN-to-SNN-Conversion Methodology}
In essence, ADMM optimization can be applied in both SNN compression and ANN compression, and it is interesting to compare their compression performance. In this subsection, we use an extra experiment to do a simple analysis. We compare with the ANN-to-SNN-conversion methodology on MNIST. As given in Table \ref{tab:comparison_converted_SNN}, the direct compression of the SNN model using our methodology with 25\% connection sparsity ($s=25\%$) and 3-bit weight bitwidth ($b=3$) can achieve 98.59\% accuracy. In contrast, although the compressed ANN can achieve higher accuracy than our compressed SNN, the resulting converted SNN will lose accuracy during the model conversion if $T$ is not large enough. Usually, in order to maintain accuracy, the value of $T$ in the ANN-to-SNN-conversion methodology needs to be tens of times larger than our $T=10$.

\begin{table}[!htbp]
\caption{ Accuracy comparison between the ANN-to-SNN-conversion methodology and our direct training methodology on MNIST. }
\vspace{3pt}
\centering
\label{tab:comparison_converted_SNN}
\renewcommand\arraystretch{1.1}
  \resizebox{0.485\textwidth}{!}{
\begin{tabular}{ccccccc}
\hline\hline
\multicolumn{7}{c}{Compressed SNN ($s=25\%,~b=3,~T=10$): 98.59\%} \\ \hline
\multicolumn{7}{c}{Compressed ANN ($s=25\%,~b=3$): 98.84\%}  \\ \hline
\#timestep ($T$) & 10 & 50 & 100 & 250 & 500 & 1000 \\
Converted SNN & 11.35\% & 48.51\% & 80.14\% & 98.25\% & 98.75\% & 98.79\% \\
\hline\hline
\end{tabular}} 
\end{table}

\section{Conclusion and Discussion}\label{sec:conclusion}

In this paper, we combine STBP and ADMM to compress SNN models in two aspects: connection pruning and weight quantization, which greatly shrinks the memory space and baseline operations. Furthermore, we propose activity regularization to lower the number of dynamic spikes, which reduces the active operations. The three compression approaches can be used in a single paradigm or a joint paradigm according to actual needs. Our solution is the first work that investigates the SNN compression problem in a comprehensive manner by exploiting all compressible components and defining quantitative evaluation metrics. We demonstrate much better compression performance than prior work. 

Through extensive contrast experiments along with in-depth analyses, we observe several interesting insights for SNN compression. First, the weight quantization contributes most to the memory reduction (i.e. $R_{mem}$) while the activity regularization contributes most to the spike reduction (i.e. $R_s$). Second, there is a trade-off between $R_{mem}$ and $R_s$, and $R_{ops}$ representing the overall compression ratio could approximately reflect the accuracy after compression. Third, the gentle compression of multi-way information usually pays less accuracy loss than the aggressive compression of only single-way information. Therefore, we recommend the joint-way compression if we expect a better overall compression performance. Fourth, we observe that SNN models show a good tolerance to the weight quantization. Finally, the accuracy drop of SNNs on event-driven datasets (e.g. N-MNIST) is higher than that on static image datasets (e.g. MNIST) under low compression ratios but quite graceful when coming to aggressive compression. These observations will be important to determine the best compression strategy in real-world applications with SNNs.

Although we provide a powerful solution for comprehensive SNN compression, there are still several issues that deserve investigations in future work. We focus more on presenting our methodology and just give limited testing results due to the tight budgets on page and time. This is acceptable for a starting work to study comprehensive SNN compression; whereas, in order to thoroughly understand the joint-way compression and mine more insights, a wider range of experiments (e.g. with different settings of compression hyper-parameters, on different benchmarking datasets, using more intuitive visualizations, etc.) are highly demanded. Reinforcement learning might be a promising tool to search the optimal compression strategy \cite{he2018amc, wang2019haq} if substantial resources are available. For simplicity, we just focus on the element-wise sparsity with an irregular pattern that impedes efficient running due to the large indexing overhead. The structured sparsity \cite{zhang2018structadmm} seems helpful to optimize the execution performance. Incorporating the hardware architecture constraints into the design of compression algorithm should be considered to achieve practical saving of latency and energy on neuromorphic devices.  In addition, the comparison of the compression effects on ANNs and SNNs is also an interesting topic.

\bibliography{./ref/main.bib}

\vspace{-30pt}

\begin{IEEEbiography}[{\includegraphics[width=1in, height=1.25in, clip, keepaspectratio]{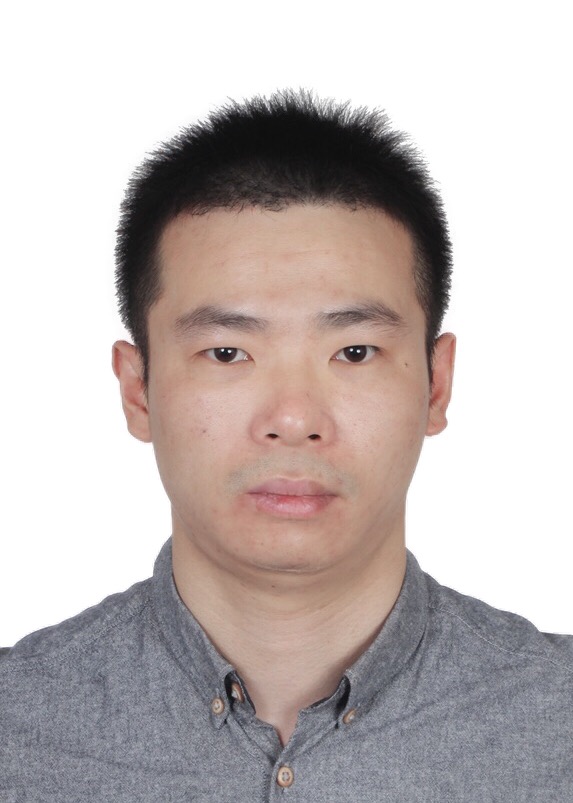}}] {Lei Deng} received the B.E. degree from University of Science and Technology of China, Hefei, China in 2012, and the Ph.D. degree from Tsinghua University, Beijing, China in 2017. He is currently a Postdoctoral Fellow at the Department of Electrical and Computer Engineering, University of California, Santa Barbara, CA, USA. His research interests span the areas of brain-inspired computing, machine learning, neuromorphic chip, computer architecture, tensor analysis, and complex networks. Dr. Deng has authored or co-authored over 50 refereed publications. He was a PC member for \emph{International Symposium on Neural Networks (ISNN)} 2019. He currently serves as a Guest Associate Editor for \emph{Frontiers in Neuroscience} and \emph{Frontiers in Computational Neuroscience}, and a reviewer for a number of journals and conferences. He was a recipient of MIT Technology Review Innovators Under 35 China 2019.
\end{IEEEbiography}
\vspace{-30pt}

\begin{IEEEbiography}[{\includegraphics[width=1in, height=1.25in, clip, keepaspectratio]{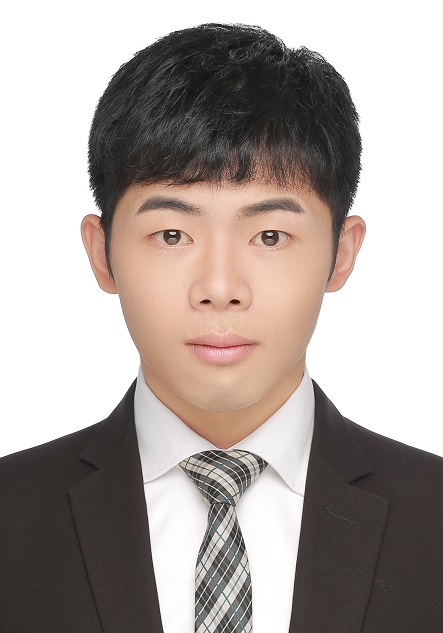}}] {Yujie Wu} received the B.E. degree in Mathematics and Statistics from Lanzhou University, Lanzhou, China in 2016. He is currently pursuing the Ph.D. degree at the  Center  for  Brain Inspired  Computing  Research  (CBICR), Department of Precision Instrument, Tsinghua University, Beijing, China. His current research interests include spiking neural networks, neuromorphic device, and brain-inspired computing.
\end{IEEEbiography}
\vspace{-30pt}

\begin{IEEEbiography}[{\includegraphics[width=1in, height=1.25in, clip, keepaspectratio]{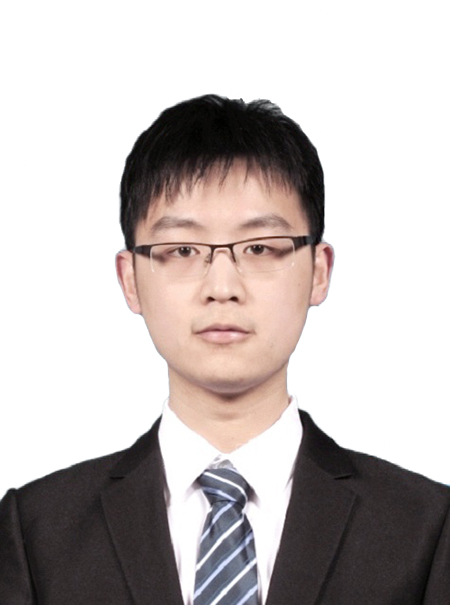}}] {Yifan Hu} received the B.S. degree from Tsinghua University, Beijing, China in 2019. He is currently pursuing the Ph.D. degree at the Center for Brain Inspired Computing Research (CBICR), Department of Precision Instrument, Tsinghua Unviersity, Beijing, China. His current research interests include deep learning and neuromorphic computing.
\end{IEEEbiography}
\vspace{-30pt}

\begin{IEEEbiography}[{\includegraphics[width=1in,height=1.25in,clip,keepaspectratio]{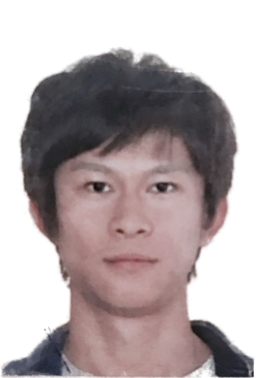}}] {Ling Liang} received the B.E. degree from Beijing University of Posts and Telecommunications, Beijing, China in 2015, and M.S. degree from University of Southern California, CA, USA in 2017. He is currently pursuing the Ph.D. degree at Department of Electrical and Computer Engineering,  University of California, Santa Barbara, CA, USA. His current research interests include machine learning security, tensor computing, and computer architecture.
\end{IEEEbiography}
\vspace{-30pt}

\begin{IEEEbiography}[{\includegraphics[width=1in,height=1.25in,clip,keepaspectratio]{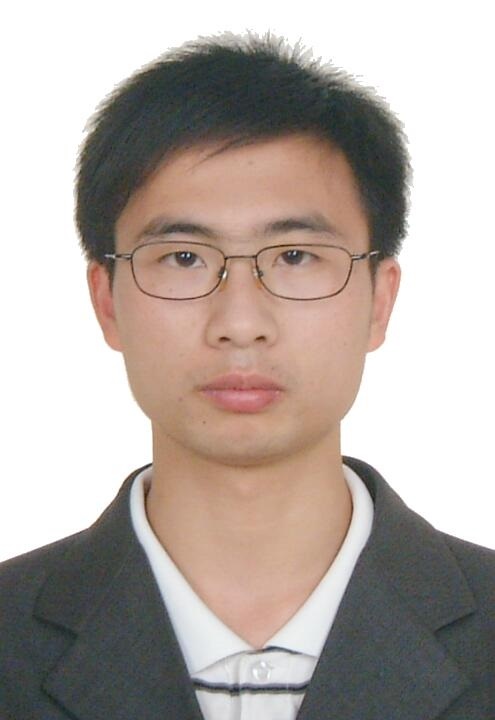}}]{Guoqi Li}  received the B.E. degree from the Xi’an University of Technology, Xi’an, China in 2004, the M.E. degree from Xi’an Jiaotong University, Xi’an, China in 2007, and the Ph.D. degree from Nanyang Technological University, Singapore, in 2011. He was a Scientist with Data Storage Institute and the Institute of High Performance Computing, Agency for Science, Technology and Research (ASTAR), Singapore from 2011 to 2014. He is currently an Associate Professor with Center for Brain Inspired Computing Research (CBICR), Tsinghua University, Beijing, China. His current research interests include machine learning, brain-inspired computing, neuromorphic chip, complex systems and system identification. Dr. Li is an Editorial-Board Member for \emph{Control and Decision} and an Associate Editor for \emph{Frontiers in Neuroscience, Neuromorphic Engineering}. He was the recipient of the 2018 First Class Prize in Science and Technology of the Chinese Institute of Command and Control, Best Paper Awards (\emph{EAIS} 2012 and \emph{NVMTS} 2015), and the 2018 Excellent Young Talent Award of Beijing Natural Science Foundation. 
\end{IEEEbiography}
\vspace{-30pt}

\begin{IEEEbiography}[{\includegraphics[width=1in,height=1.25in,clip,keepaspectratio]{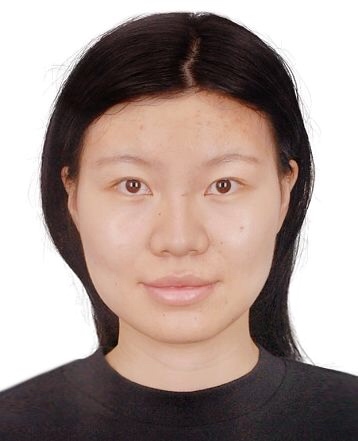}}] {Xing Hu} received the B.S. degree from Huazhong University of Science and Technology, Wuhan, China, and Ph.D. degree from University of Chinese Academy of Sciences, Beijing, China in 2009 and 2014, respectively. She is currently a Postdoctoral Fellow at the Department of Electrical and Computer Engineering, University of California, Santa Barbara, CA, USA. Her current research interests include emerging memory system, domain-specific hardware, and machine learning security.
\end{IEEEbiography}
\vspace{-30pt}

\begin{IEEEbiography}[{\includegraphics[width=1in, height=1.25in, clip, keepaspectratio]{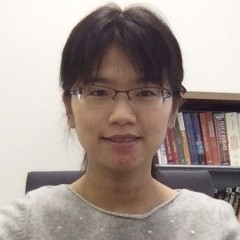}}] {Yufei Ding} received her B.S. degree in Physics from University of Science and Technology of China, Hefei, China in 2009, M.S. degree from The College of William and Mary, VA, USA in 2011, and the Ph.D. degree in Computer Science from North Carolina State University, NC, USA in 2017. She joined the Department of Computer Science, University of California, Santa Barbara as an Assistant Professor since 2017. Her research interest resides at the intersection of Compiler Technology and (Big) Data Analytics, with a focus on enabling High-Level Program Optimizations for data analytics and other data-intensive applications. She was the receipt of NCSU Computer Science Outstanding Research Award in 2016 and Computer Science Outstanding Dissertation Award in 2018.
\end{IEEEbiography}
\vspace{-30pt}

\begin{IEEEbiography}[{\includegraphics[width=1in, height=1.25in, clip, keepaspectratio]{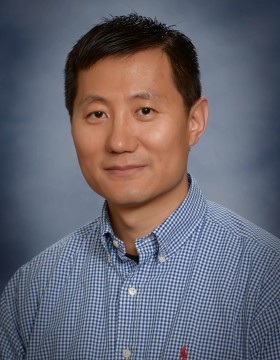}}] {Peng Li} received the Ph.D. degree in electrical and computer engineering from Carnegie Mellon University, Pittsburgh, PA, USA in 2003. He was a Professor with Department of Electrical and Computer Engineering, Texas A\&M University, College Station, TX, USA from 2004 to 2019. He is presently a Professor with Department of Electrical and Computer Engineering, University of California, Santa Barbara, CA, USA. His research interests include integrated circuits and systems, computer-aided design, brain-inspired computing, and computational brain modeling.

His work has been recognized by various distinctions including the ICCAD Ten Year Retrospective Most Influential Paper Award, four IEEE/ACM Design Automation Conference Best Paper Awards, the IEEE/ACM William J. McCalla ICCAD Best Paper Award, the ISCAS Honorary Mention Best Paper Award from the Neural Systems and Applications Technical Committee of IEEE Circuits and Systems Society, the US National Science Foundation CAREER Award, two Inventor Recognition Awards from Microelectronics Advanced Research Corporation, two Semiconductor Research Corporation Inventor Recognition Awards, the William and Montine P. Head Fellow Award and TEES Fellow Award from the College of Engineering, Texas A\&M University. He was an Associate Editor for \emph{IEEE Transactions on Computer-Aided Design of Integrated Circuits and Systems} from 2008 to 2013 and \emph{IEEE Transactions on Circuits and Systems-II: Express Briefs} from 2008 to 2016, and he is currently a Guest Associate Editor for \emph{Frontiers in Neuroscience}. He was the Vice President for Technical Activities of IEEE Council on Electronic Design Automation from 2016 to 2017.
\end{IEEEbiography}
\vspace{-30pt}

\begin{IEEEbiography}[{\includegraphics[width=1in,height=1.25in,clip,keepaspectratio]{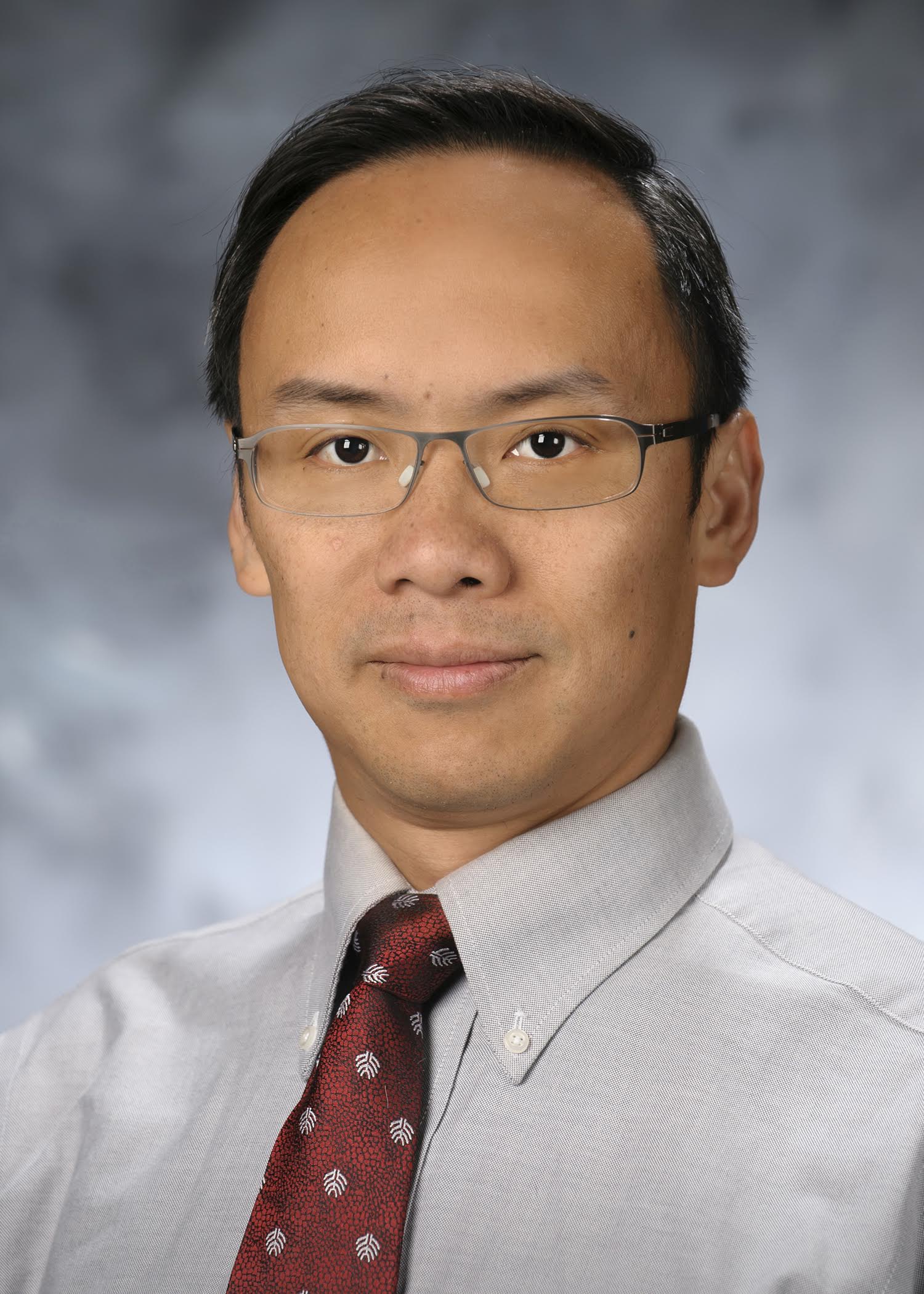}}]{Yuan Xie}  received the B.S. degree in Electronic Engineering from Tsinghua University, Beijing, China in 1997, and M.S. and Ph.D. degrees in Electrical Engineering from Princeton University, NJ, USA in 1999 and 2002, respectively. He was an Advisory Engineer with IBM Microelectronic Division, Burlington, NJ, USA from 2002 to 2003. He was a Full Professor with Pennsylvania State University, PA, USA from 2003 to 2014. He was a Visiting Researcher with Interuniversity Microelectronics Centre (IMEC), Leuven, Belgium from 2005 to 2007 and in 2010. He was a Senior Manager and Principal Researcher with AMD Research China Lab, Beijing, China from 2012 to 2013. He is currently a Professor with the Department of Electrical and Computer Engineering, University of California at Santa Barbara, CA, USA. His interests include VLSI design, Electronics Design Automation (EDA), computer architecture, and embedded systems. 

Dr. Xie is an expert in computer architecture who has been inducted to \emph{ISCA}/\emph{MICRO}/\emph{HPCA} Hall of Fame and IEEE/AAAS/ACM Fellow. He was a recipient of the 2020 IEEE Computer Society Edward J. McCluskey Technical Achievement Award, 10-Year Retrospective Most Influential Paper Award (\emph{ASPDAC} 2019), Best Paper Awards (\emph{HPCA} 2015, \emph{ICCAD} 2014, \emph{GLSVLSI} 2014, \emph{ISVLSI} 2012, \emph{ISLPED} 2011, \emph{ASPDAC} 2008, \emph{ASICON} 2001) and Best Paper Nominations (\emph{ASPDAC} 2014, \emph{MICRO} 2013, \emph{DATE} 2013, \emph{ASPDAC} 2010/2009, \emph{ICCAD} 2006), the 2016 IEEE Micro Top Picks Award, the 2008 IBM Faculty Award, and the 2006 NSF CAREER Award. He served as the TPC Chair for \emph{ICCAD} 2019, \emph{HPCA} 2018, \emph{ASPDAC} 2013, \emph{ISLPED} 2013, and \emph{MPSOC} 2011, a committee member in IEEE Design Automation Technical Committee (DATC), the Editor-in-Chief for \emph{ACM Journal on Emerging Technologies in Computing Systems}, and an Associate Editor for \emph{ACM Transactions on Design Automations for Electronics Systems}, \emph{IEEE Transactions on Computers}, \emph{IEEE Transactions on Computer-Aided Design of Integrated Circuits and Systems}, \emph{IEEE Transactions on VLSI, IEEE Design and Test of Computers}, and \emph{IET Computers and Design Techniques}. Through extensive collaboration with industry partners (e.g. AMD, HP, Honda, IBM, Intel, Google, Samsung, IMEC, Qualcomm, Alibaba, Seagate, Toyota, etc.), he has helped the transition of research ideas to industry. 
\end{IEEEbiography}

\end{document}